\documentclass{article} 
\usepackage{conference,times}

\usepackage{amsmath,amsfonts,bm}

\def\eqref#1{equation~\ref{#1}}

\def\1{\bm{1}}

\DeclareMathAlphabet{\mathsfit}{\encodingdefault}{\sfdefault}{m}{sl}
\SetMathAlphabet{\mathsfit}{bold}{\encodingdefault}{\sfdefault}{bx}{n}

\usepackage{wrapfig}
\usepackage{hyperref}
\usepackage{url}
\usepackage{graphicx}
\usepackage{booktabs}
\usepackage{amssymb}
\usepackage{pifont}
\newcommand{\cmark}{\ding{51}}%
\newcommand{\xmark}{\ding{55}}%
\usepackage{colortbl}

\usepackage{amsmath}
\newcommand{\modelname}{\emph{DiRL}}
\newcommand{\modelnameplain}{DiRL}
\newcommand{\modelnamenossl}{diversity-inducing}
\usepackage{latexsym}
\usepackage{arydshln}

\usepackage{multirow}
\definecolor{Gray}{gray}{0.9}

\title{Attention De-sparsification Matters: Inducing Diversity in Digital Pathology Representation Learning}

\author{Saarthak Kapse$^1$, Srijan Das$^2$, Jingwei Zhang$^1$, Rajarsi R. Gupta$^1$, Joel Saltz$^1$,  \textbf{Dimitris Samaras}$^1$, \\ \textbf{Prateek Prasanna}$^1$ \\
$^1$Stony Brook University, $^2$UNC Charlotte\\
}

\begin{document}

\maketitle
\begin{abstract}
We propose \modelname, a \textbf{D}iversity-\textbf{i}nducing \textbf{R}epresentation \textbf{L}earning technique for histopathology imaging. Self-supervised learning techniques, such as contrastive and non-contrastive approaches, have been shown to learn rich and effective representations of digitized tissue samples with limited pathologist supervision. 
Our analysis of vanilla SSL-pretrained models' attention distribution reveals an insightful observation: \textit{sparsity in attention}, i.e, models tends to localize most of their attention to some prominent patterns in the image. Although attention sparsity can be beneficial in natural images due to these prominent patterns being the object of interest itself, this can be sub-optimal in digital pathology; this is because, unlike natural images, digital pathology scans are not object-centric, but rather a complex phenotype of various spatially intermixed biological components. Inadequate diversification of attention in these complex images could result in crucial information loss. To address this, we leverage cell segmentation to densely extract multiple histopathology-specific representations, and then propose a prior-guided dense pretext task for SSL, designed to match the multiple corresponding representations between the views. Through this, the model learns to attend to various components more closely and evenly, thus inducing adequate diversification in attention for capturing context rich representations. Through quantitative and qualitative analysis on multiple tasks across cancer types, we demonstrate the efficacy of our method and observe that the attention is more globally distributed.
\end{abstract}


\section{Introduction}

Computational pathology is a rapidly emerging field that aims at analyzing high resolution images of biopsied or resected tissue samples. Advancements in computer vision and deep learning has enabled learning of the rich phenotypic information from whole slides images (WSIs) to understand mechanisms contributing to disease progression and patient outcomes. 
Acquiring crop-level localized annotations for WSIs is expensive and often not feasible; only slide-level pathologist labels are usually available. In such scenarios, weak supervision is a commonly utilized strategy, where crops are \textit{embedded into representations} in the first stage, followed by considering these WSI-crops' representation as a bag for multiple instance learning (MIL). Now the question remains, \textit{how do we learn a model to effectively encode the crops into rich representations?} Traditionally, ImageNet \citep{imagenet} pretrained neural networks are utilized to extract the representations \citep{CLAM, sparseconvmil, Transmil}. However ImageNet and pathology datasets are composed of different semantics; while the former contains object-centric natural images, the later consists of images with spatially distributed biological components such as cells, glands, stroma, etc. Therefore, to learn domain-specific features of WSI-crops in the absence of localized annotations, various self-supervised learning (SSL) techniques are recently gaining traction \citep{ciga, stacke2021learning, boyd2021self}. There studies have shown the effectiveness of models pretrained through SSL on histopathology images in downstream classification tasks when compared to model trained on ImageNet.

\begin{figure*}[t]
    \centering
    \includegraphics[width=1\linewidth]{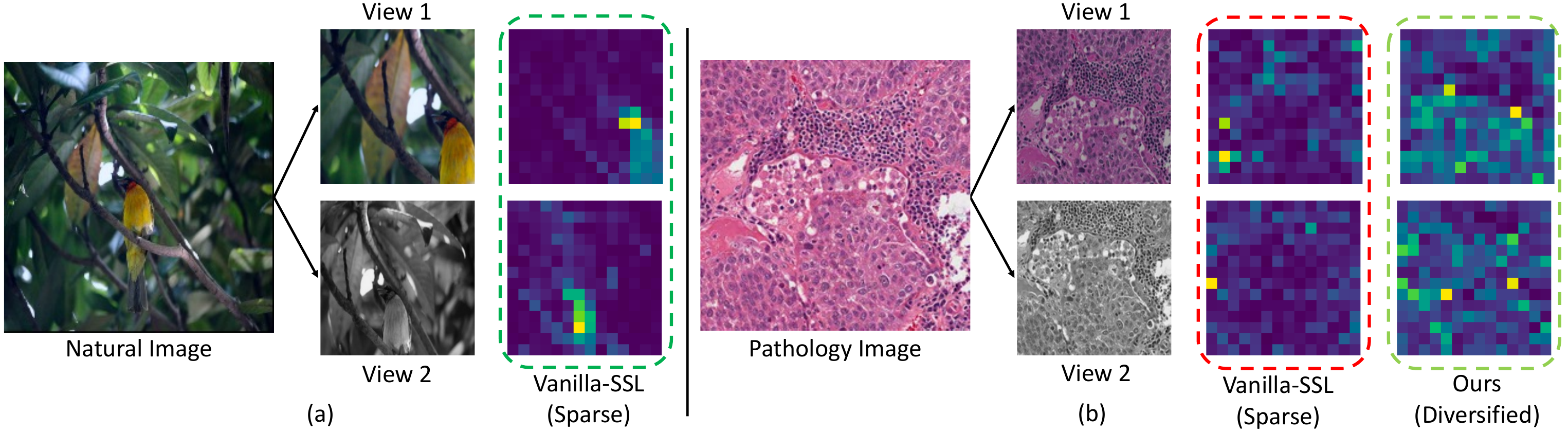}
    \caption{\textbf{Diversification of attention for encoding dense information in digital pathology.} View 1 and View 2 are two augmented views of the input image. a) Illustration of attention map from a model pretrained on ImageNet using vanilla SSL. b) Attention map of model pretrained on histopathology dataset with vanilla SSL, and with our proposed 
    pre-training strategy. In both natural imaging and digital pathology, vanilla SSL pre-training creates \textit{sparse} attention maps, i.e., it attends largely to only some prominent patterns. Although attention sparsification can be beneficial in natural image tasks such as object classification, this could be sub-optimal for encoding representations in digital pathology as it leads to loss of important contextual information.
    Through a more diversified attention mechanism, ~\modelname~ encodes dense information critical to non object-centric tasks. 
    }%
    \label{fig:motivation}%
\end{figure*}

To further analyze the role of SSL in computational pathology, we pretrained a vision transformer \citep{vit} on various WSI datasets using vanilla SSL~\citep{dino}. An in-depth analysis of the pretrained models' attention maps on WSI-crops led us to a striking observation: \textbf{sparsity in attention maps}. The model tends to localize most of its attention to a small fraction of regions, leading to sub-optimal representation learning. To further validate our observation, we visualized the attention maps of a self-supervised ImageNet pretrained model on natural images (see Fig. \ref{fig:motivation}). Similar observations led us to conclude that this is a property of SSL rather than of data.  We believe that sparsity in attention might potentially benefit the performance in some natural imaging tasks such as object classification. This stems from the fact that during SSL, the model is tasked to match the two views, optimizing which leads the model to focus on the prominent patterns. For example, in Fig.~\ref{fig:motivation}(a), for an object-centric ImageNet example, since the prominent pattern is the object (eg. bird) itself \citep{selfpatch}, the model tends to center its attention towards the object, thus benefiting numerous downstream applications (for eg., bird classification). In contrast, WSI-crops are \textit{not object-centric}, rather they constitute a \textit{spatial distribution of complex structures such as cells, glands, their clusters and organizations, etc}, see Fig.~\ref{fig:motivation}(b). Encoding this dense information available into a holistic representation demands the model to focus more diversely to various histopathology primitives and not just to specific ones. Conversely, the vanilla SSL model pretrained on histopathology only sparsely attends to the important regions (Fig.~\ref{fig:motivation}(b)), i.e., there is inadequate diversity in attention. We \textit{hypothesize that this sparsely attending model could result in encoding sub-optimal representations, as fine-grained context-rich details are often ignored}. 

To address this issue of inadequate attention diversity, we propose \textbf{~\modelname}, a diversity-inducing 
pre-training technique, tailored to enhance representation learning in digital pathology. Each WSI-crop consists of \textit{two regions: cellular regions (one containing cells) and non-cellular regions (containing no cells)}. We leverage an off-the-shelf cell segmentation pipeline to identify these regions. This domain-specific knowledge is then utilized to extract \textbf{region-level representations} separately for the cellular and non-cellular regions. We further propose to encode the inter- and intra-spatial interplay of two regions. This biologically-inspired step \citep{saltz2018spatial, fassler2022spatial} is achieved through a transformer-based disentangle block to encode the self-interaction within the regions, and cross-interaction between both the regions, termed as \textbf{disentangled representations}. In contrast to vanilla SSL frameworks that leverage one image-level representation for a WSI-crop, 
our prior-guided representation learning framework leverages histology-specific domain knowledge to densely extract a set of region-level and disentangled representations.
We then task our framework to match all the corresponding representations between the views. We hypothesize that \textit{optimizing this dense matching objective between the views would encourage the model to diversify its attention to various regions; matching assorted representations would then enforce the model to explore diverse image-regions relevant for each such representations.} 
We validate this hypothesis through consistent improvements in performance on multiple downstream tasks such as slide-level and patch-level classifications. Our qualitative analysis on attention distribution of the
pretrained models reveals that our \modelname~framework can effectively de-sparsify attention, thereby learning global context-rich representations, unlike existing methods.
To summarize our main contributions, we:

\begin{itemize} 
    \item Demonstrate that attention sparsification in self-supervised learning may lead to learning sub-optimal representations in digital pathology classification tasks.
    \item Design a novel domain-aware pretext task to de-sparsify attention maps and achieve enhanced 
    representations for digital pathology.
    \item Demonstrate the efficacy of our \textit{DiRL}
    through slide-level and patch-level classification tasks on three WSI datasets and two patch datasets.
\end{itemize}

\section{Related Work}

In this section, we briefly discuss vision transformers, SSL and its dense counterpart, and their application in computational pathology. \\
\textbf{Vision transformers.} Inspired by the success of self-attention modules in language models \citep{attention}, vision transformers (ViTs) \citet{vit, swin, deit, groupvit, xcit, maxvit} have been have been proposed to exploit non-local spatial dependencies in the imaging domain. 
Recent studies \citet{transpath, chen2022self, hipt, scorenet, gao2021instance, gashis} have demonstrated the promise of transformer-based architectures in modeling histopathology imaging for cancer diagnosis and prognosis. However, to the best of our knowledge, no existing work has leveraged the flexibility of attention mechanism in transformers to instill the biology-relevant domain knowledge into vision transformers. For example, interaction between concepts/primitives such as tumor nuclei and stroma or between lymphocytic cells plays an important role in disease pathophysiology and treatment outcome. Our proposed method takes a step in this direction through a domain-driven pretext task.\\ 
\textbf{Image-level SSL} aims at learning visual representations through different pretext tasks. Contrastive and non-contrastive methods such as \citet{simclr, moco, dino}, have shown tremendous potential in learning robust and rich representation in natural imaging. Building upon them, studies such as \citet{ciga, stacke2021learning, DSMIL, cdnet, boyd2021self, chen2022self, kurian2022improved} have explored SSL pre-training in histopathlogy image analysis. \\
\textbf{Region-level SSL} aims to further boost information encoding through dense pre-training techniques such as \citet{esvit, selfpatch, densecl}. These techniques impose additional constraints to match 1) region-level correspondences across the two views of the data or 2) neighbor-level intra-view correspondences within the data. 
Studies such as \citet{wen2022self, yang2022concl, henaff2021efficient} have explored utilizing segmentation-based or clustering-based regions in self-supervision to enhance representation learning. However, the goal of these studies is to mainly improve the transfer performance for dense-prediction tasks such as \textit{object detection} and \textit{segmentation}. In contrast, we tailored a dense pre-training strategy in histopathology to enforce the model to focus on diverse-regions thus diversifying model's attention. This diversified attention encourages the model in effectively encoding the complex information about various histology components, thereby augmenting \textit{classification} performance.

\section{Methodology}

\begin{figure*}[t]
    \centering
    \includegraphics[width=1\linewidth]{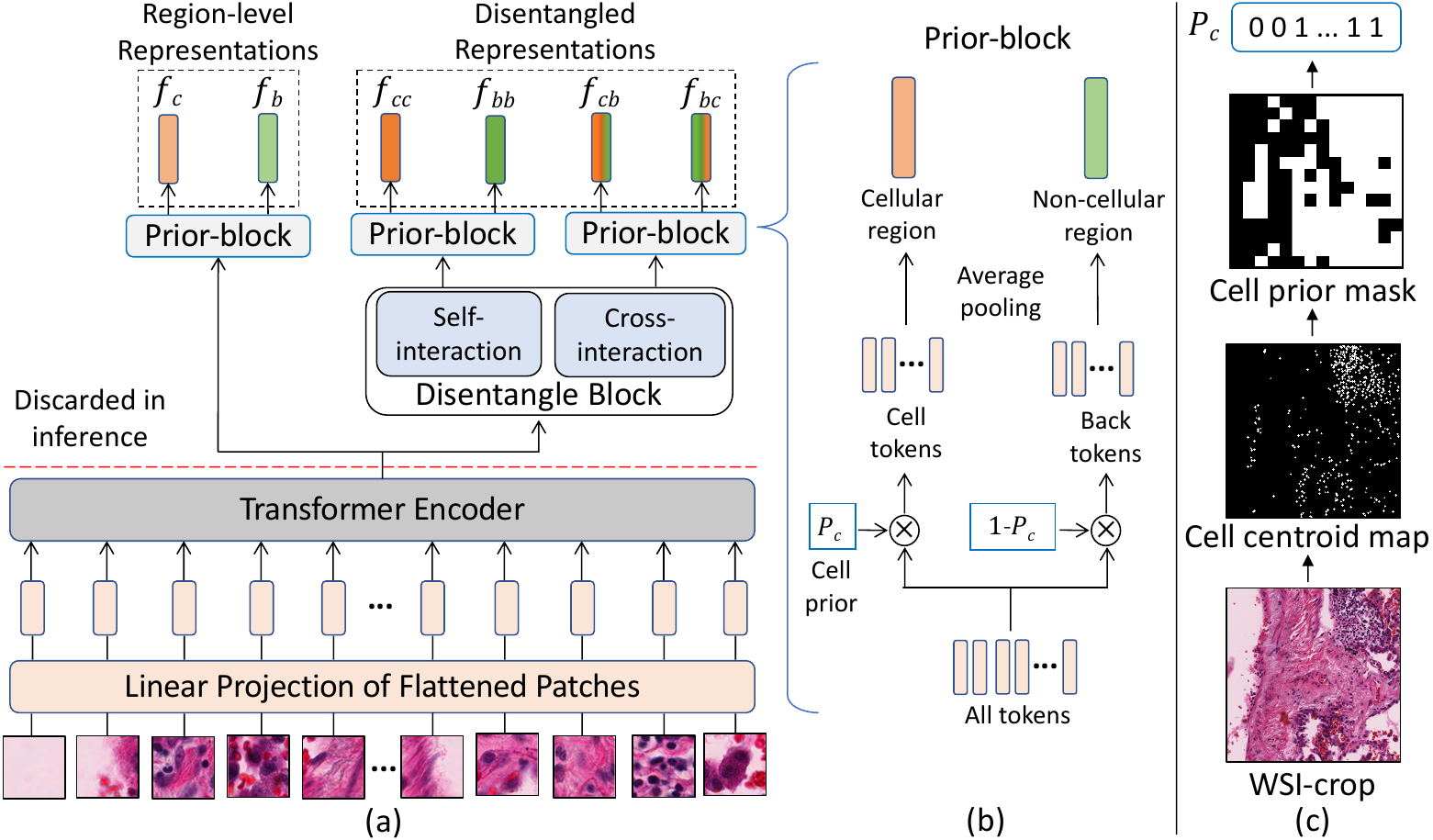}
    \caption{\textbf{Overview of the proposed \modelname~framework:} a) A WSI-crop is patchified and fed into a linear projection layer followed by a transformer encoder. The output is fed to a Prior-block and to a Disentangle block. The Prior-block pools region-level representations separately for cellular and non-cellular regions. The Disentangle block encodes spatial interplay between the two regions followed by prior-block to extract region-level disentangled representations. b) Cell priors $P_c$ and $1-P_c$ pool the tokens associated with the cellular and the non-cellular regions, respectively, followed by average pooling to extract region-level features. c) Cell segmentation from WSI-crop followed by extraction of cell centroid map. Cell prior mask is generated by discretizing the cell centroid map into patches. A Cell prior vector $P_c$ is then produced from the cell prior mask. After
    pre-training, modules above the red dashed line are discarded in the inference stage.}
    \label{fig:framework}
\end{figure*}

In this section, we first describe a na\"{\i}ve vision transformer framework for Whole Slide Images (WSIs). This is followed by explaining how cell segmentation can be used as a prior in pre-training for WSIs. Next, we present the extraction of region-specific representations using our proposed cell-back pooling and disentangle block. Finally, we present \modelname, our diversity-inducing 
pre-training technique, to learn discriminative features for WSI patches which are subsequently leveraged by a multiple instance learning (MIL) framework for downstream classification tasks. An overview of the proposed architecture is shown in Fig.~\ref{fig:framework}(a).

\noindent\textbf{Preliminaries.} For an understanding of the primary components in a transformer such as MSA (Multi-head Self-Attention), LN (Layer Normalization), and MLP (Multi-Layer Perceptron), we refer the readers to~\citet{attention}.

\subsection{Vision Transformer for WSI}

\begin{figure*}[t]
    \centering
    \includegraphics[width=1\linewidth]{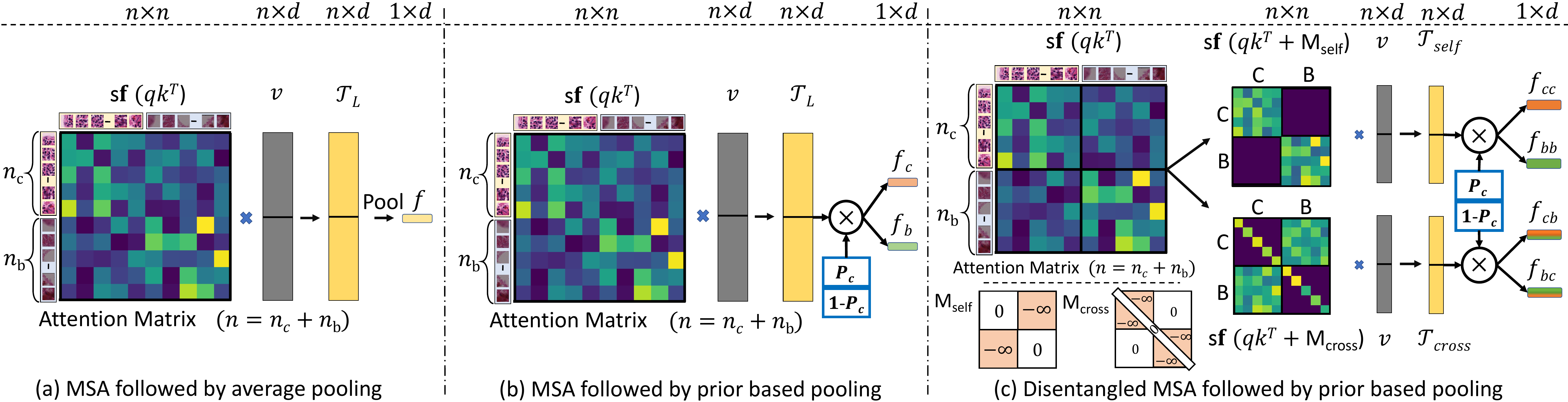} 
    \caption{a) Illustration of the $n \times n$ attention matrix from the last layer of the transformer encoder, where $q, k, v$ are projections of tokens in transformer block. Matrix multiplication of attention matrix with value $v$, followed by average pooling across all the tokens generates the representation ($f$) for $w_i$. b) Cell prior $P_c$ is utilized to separately pool cell tokens and back tokens to extract region-level representations. c) Tokens from the $L^{th}$ layer are interacted in the transformer-based disentangle block, forming the attention matrix. Attention masks $M_{self}$ and $M_{cross}$ are added with the attention matrix, generating desired matrices for disentanglement. Note that $\mathrm{sf}$ denotes $\mathrm{softmax}$ activation. Matrices are then multiplied with $v$ followed by prior-based pooling, thus extracting four representations encoding spatial interplay in $w_i$. For clarity, the cellular and non-cellular region patches ($n_c$ and $n_b$ respectively) are arranged in separate groups.}%
    \label{fig:attention}%
\end{figure*}

From each WSI, $\mathcal{W}$, $w_1$, $w_2$, \dots $w_N$ crops are extracted, where $N$ is variable for each $\mathcal{W}$. Each $w_i$ is then decomposed into $n$ patches $\mathcal{X} = [X^1, X^2, ..., X^n] \in \mathcal{R}^{n\times p\times p\times 3}$, where ($p$, $p$) is the spatial size of each patch. Each patch is transformed into a token using a shared linear projection layer, 
\begin{equation} \label{patchembed}
\mathcal{T}_{0} = [X^1\textbf{E}; X^2\textbf{E};...,X^n\textbf{E}]
\end{equation}
where $\textbf{E}$ are convolutional filters operating on each patch with $d$ number of $p \times p$ size filters, thus extracting a $d$ dimensional feature vector for patch. 
This is followed by adding 1D learnable position embedding as in~\citet{attention}. The transformer block models the relationship between the tokens using a multi-head self-attention block:
\begin{equation}
  \mathcal{T}^{'}_l = \mathcal{T}_{l-1} + \mathrm{MSA}(\mathrm{LN}(\mathcal{T}_{l-1})); \hspace{0.2in} \mathcal{T}_l = \mathcal{T}^{'}_{l} + \mathrm{MLP}(\mathrm{LN}(\mathcal{T}^{'}_{l}))
\end{equation}
where $l$ is index of the $l^{th}$ block of transformer encoder, composed of $L$ stacked transformer blocks. Thus in each block, these tokens interact with each other to learn representations for each $w_i$. The resulting $\mathcal{T}_{L}$ of dimension ($n$, $d$) is average pooled across all the $n$ tokens to compute the image-level representation $f$ of dimension ($1$, $d$) as shown in Fig.~\ref{fig:attention}(a). 


\subsection{Cell segmentation as domain prior} \label{Cell segmentation as domain prior}

Each WSI-crop $w_i$ consists of \textbf{two regions}, one containing cells and the other without cells. There has been substantial advancements in deep learning research pertaining to cell segmentation; this stems from the important role of image analysis and machine learning algorithms in visual interpretation of cellular biology (morphology and spatial arrangement) in digitized pathology scans \citep{lu2021feature,shaban2022digital,ding2022image}. Identifying the cellular and non-cellular regions in $w_i$ can be achieved by exploiting the cell segmentation output as prior via techniques such as \citet{sahasrabudhe2020self, hou2020dataset, vahadane2013towards}. Following cell segmentation, the centroids are extracted to yield the cell centroid map, a binary map ($\mathcal{C}$) of values zeros and ones, with $\mathcal{C}_{i,j} = 1$ if centroid of any cell is present at ($i,j$) pixel in WSI-crop. We term this as cell prior mask as shown in Fig.~\ref{fig:framework}(c). Since cell segmentation is routinely used in computational pathology, we use off-the-shelf, well established cell segmentation pipelines instead of training a new model. To be coherent with ViT, $\mathcal{C}$ is decomposed into $n$ patches $\mathcal{C} = [C^1, C^2, ..., C^n]$ of size ($p, p$). Each patch is transformed as follows: $C^i = \mathrm{MaxPool}(C^i)$, i.e., if the patch $C^i$ contains one or more centroids, it becomes one, else it remains zero.  
Thus, the cell prior mask is downsampled into a binary vector of dimension ($n$, $1$), denoting the presence of cells in each patch of $w_i$. We term this vector as cell prior ($P_c$), which is invoked to extract the region-specific representations for each $w_i$.

\subsection{Prior-block for Cell-Back pooling}\label{sec_cellback}
Following $L$ stacks of transformer blocks, a set of tokens $\mathcal{T}_{L} \in \mathcal{R}^{n \times d}$ is fed to a prior-block. In this block, the tokens can be categorized into (a) \textit{cell} tokens, implying the tokens whose input patches contain at least a cell and (b) background or \textit{back} tokens, whose input patches do not capture any cells. 

The cell and back tokens are separately encoded to represent region-level features from the cell prior $P_c$ as follows:  
\begin{equation} \label{equation3}
  f_c = \frac{P_c^T\mathcal{T}_{L}}{\sum P_c}; \hspace{0.2in} f_b = \frac{(1-P_c)^T\mathcal{T}_{L}}{\sum 1-P_c} 
\end{equation}

$f_c$ is the average pooled representation of all the cell tokens, i.e. representation of the cellular region. $f_b$ is the average pooled representation of all the back tokens, representing non-cellular regions. In the process, \textit{Cell-Back Pooling} is exploited in the prior-block to extract two \textbf{region-level representations} as shown in Fig.~\ref{fig:attention}(b).

\subsection{Disentangle block}\label{sec_disentangle}

We take a step towards obtaining region-level representation by proposing a transformer block for disentangling the cellular and non-cellular regions. This disentanglement is performed to encode self-interaction in each region and cross-interaction between the two regions. To accomplish this disentanglement, we devise two attention masks, $M_{self}$ and $M_{cross}$, each of dimension ($n \times n$) as shown in Fig.~\ref{fig:attention}(c). The goal of $M_{self}$ is to only allow token interaction between the same regions, i.e.,  cell-cell and back-back. In contrast, $M_{cross}$ allows tokens to interact between the different regions (cell-back). The masks $M_{self}$ and $M_{cross}$ are computed as:

\begin{equation}
\begin{aligned}
   & M_{self}(i,j) = \left \{
  \begin{aligned}
    &0, && \text{if}\ P_c(i) = P_c(j) \\
    &-\infty, && \text{otherwise}
  \end{aligned} \right.;  \\ 
    & M_{cross}(i,j) = \left \{
  \begin{aligned}
    &0, && \text{if}\ P_c(i) \neq P_c(j) \hspace{0.04in} or \hspace{0.04in} i=j \\
    &-\infty, && \text{otherwise}
  \end{aligned} \right.
\end{aligned}
\end{equation}

where indices $i,j\in \{1,2,...n\}$. The intuition behind diagonal elements in $M_{cross}$ is to ensure the preservation of each token's information when interacting with tokens from another region.
Recall that in transformers, tokens are projected into three embeddings $q$, $k$, $v$ and the output of a MSA block is computed as a weighted sum of the values $v$, where the weight assigned to each value is determined by a self-attention operation $\mathrm{softmax}(qk^T)$.
Unlike standard MSA in $w_i$, where all the tokens from cellular and non-cellular regions interact with each other through self-attention (see Fig.~\ref{fig:attention}(a)), we propose to disentangle the interactions between the regions. $M_{self}$ and $M_{cross}$ are linearly combined with the attention matrix to obtain disentangled self-attention matrices as follows:

\begin{equation}
\begin{aligned}
  \mathrm{MSA}_{self} = \mathrm{softmax}(\frac{qk^T + M_{self}}{\sqrt{d}}) \hspace{0.1cm} v; \\ \mathrm{MSA}_{cross} = \mathrm{softmax}(\frac{qk^T + M_{cross}}{\sqrt{d}}) \hspace{0.1cm} v
  \end{aligned}
\end{equation}

 Note that attention masks $M_{self}$ and $M_{cross}$ are linearly combined before the $\mathrm{softmax}$ activation to ensure that the sum of each row in the self-attention matrix remains one. For the sake of brevity, we have illustrated self-attention with just one head instead of multi-head self-attention in the above equation. However, in practice, q,k,v are split into h (number of heads) parts and self-attention is then performed on all the h parts in parallel.  
 The disentangle block operates at the output of the transformer encoder $\mathcal{T}_L$ as:

\begin{equation}
\begin{aligned}
  \mathcal{T}^{'}_{self} = \mathcal{T}_{L} + \mathrm{MSA}_{self}(\mathrm{LN}(\mathcal{T}_{L})); \\ \mathcal{T}_{self} = \mathcal{T}^{'}_{self} + \mathrm{MLP}(\mathrm{LN}(\mathcal{T}^{'}_{self}))
  \end{aligned}
\end{equation}

\begin{equation}
\begin{aligned}
  \mathcal{T}^{'}_{cross} = \mathcal{T}_{L} + \mathrm{MSA}_{cross}(\mathrm{LN}(\mathcal{T}_{L})); \\ \mathcal{T}_{cross} = \mathcal{T}^{'}_{cross} + \mathrm{MLP}(\mathrm{LN}(\mathcal{T}^{'}_{cross}))
    \end{aligned}
\end{equation}

Finally, similar to \ref{sec_cellback}, the region-level features are encoded using the cell prior $P_c$ using:

\begin{equation}
\begin{aligned}
  f_{cc} = \frac{P_c^T\mathcal{T}_{self}}{\sum P_c}; \hspace{0.1in} f_{bb} = \frac{(1-P_c)^T\mathcal{T}_{self}}{\sum 1-P_c}; \\ f_{cb} = \frac{P_c^T\mathcal{T}_{cross}}{\sum P_c}; \hspace{0.1in} f_{bc} = \frac{(1-P_c)^T\mathcal{T}_{cross}}{\sum 1-P_c}  
    \end{aligned}
\end{equation}

Thus, the prior-based pooling on $\mathcal{T}_{self}$ and $\mathcal{T}_{cross}$ results in four \textbf{disentangled representations} $f_{cc}$, $f_{bb}$, $f_{cb}$, and $f_{bc}$, encoding the spatial interplay between the cellular and non-cellular regions. Thus, for each WSI-crop $w_i$ \textit{we encode six representations: two region-level representations using cell-back pooling, and four disentangled representations using disentangle block.} Our 
prior-guided pre-training framework operates on these six representations to pretrain the model.

\subsection{Diversity-inducing pre-training for WSI}
In this section, we formulate our \modelnamenossl~representation learning
(\modelname) using a widely used SSL framework for pre-training on histopathology data: DINO~\citep{dino}. However, in practice, our pre-training technique can be integrated with any pairwise SSL framework~\citep{pairwise_learning}, as demonstrated in Appendix \ref{additional_ablation}.  
DINO consists of a student and teacher branch, where the teacher is a momentum updated version of the student, thus both having same architecture (models). Different views of the input image are fed to both the branches to encode them into image-level representations. A projection head is applied on top of these representations with  $\mathrm{softmax}$ activation. SSL is performed by matching the student's output with the teacher's probability distribution through cross-entropy loss, $\mathcal{L}^{CE}$.  

In contrast to vanilla DINO, \modelname~yields six feature vectors from each branch (see Fig.~\ref{fig:framework}). Therefore, the loss function is modified as:
\begin{equation}
  \mathcal{L}^{CE} = \lambda_1 \times (\mathcal{L}^{CE}_c + \mathcal{L}^{CE}_b) + \lambda_2 \times (\mathcal{L}^{CE}_{cc} + \mathcal{L}^{CE}_{bb} + \mathcal{L}^{CE}_{cb} + \mathcal{L}^{CE}_{bc})
\end{equation}
where $\mathcal{L}^{CE}_c$ is the cross-entropy loss between projection of representation $f_c$ of student and that of teacher branch. Likewise, the projected distribution of all other corresponding representations from student and teacher are matched. 
This linear combination of losses encourages the framework to perform a dense matching of the region-level and disentangled representations of the augmented views. Consequently, the dense matching promotes the model to globally diversify the attention map (refer to Fig~\ref{fig:motivation}, Fig~\ref{fig:additional_attention_visualization}).

We propose another variant of \modelname, without the disentangle block, i.e. similarities of only projection distribution of $f_c$ and $f_b$ are maximized between the views. We name this variant as \textit{Cellback}.
Following the pre-training, only linear projection layer, position embedding, and the transformer encoder of the teacher are retained. This pretrained ViT extracts the average pooled feature representation for all $w_i$ belonging to WSI $\mathcal{W}$, to generate feature matrix of dimension ($N$, $d$), where $N$ is variable number of WSI-crops for each $\mathcal{W}$. Note that the prior is used \textit{only} at pre-training. Finally, MIL operates over this matrix for WSI slide-level analysis, as discussed next.

\subsection{Preliminary extension to multiple cell types}
In this study, we further propose extending \textit{Cellback}~\ref{sec_cellback} to incorporate multiple cell types. Precisely, in Equation~\ref{equation3}, the cell prior ($P_c$) is replaced with multiple priors, one for each of the $j$ cell classes as follows {$P_{c_1}$, $P_{c_2}$, ... $P_{c_j}$}. Note that in $\mathcal{C}$ (please refer to ~\ref{Cell segmentation as domain prior}), a patch can contain centroids of multiple cell types. Therefore for a WSI-crop the different cell priors can be overlapping vectors, i.e. multiple priors could pool the same tokens. Background prior is conceptually the same, i.e. it pools tokens that don't contain any cells. We denote this extended version as \textit{Cellback-V2}. We use HoVer-Net for segmenting and classifying cells into 5 cell types from PanNuke. Apart from each cell type level and background level representation extraction, in this version we perform average pooling of all tokens  to extract a crop-level representation. Thus Cellback-V2 extracts a total of 6 region-level and 1 crop-level representations. All the representations are projected and matched across the views with equal weightage.

\subsection{Multiple Instance Learning for slide-level tasks}
Multiple instance learning (MIL) is widely used method in WSI slide level analysis. We refer the readers to ~\citet{ABMIL,sparseconvmil,Transmil,CLAM} for an overview. We adopted DSMIL~\citep{DSMIL} framework for this work. Following 
pre-training, the pretrained model is used to extract features for WSI-crops in $\mathcal{W}$. The MIL model takes these features as input bag, optimizing the model weights through slide-level label supervision.

\section{Experiments and Results}

Our pretrained models are evaluated for both slide-level and patch-level classification tasks. As a \textit{Baseline}, we  a vision transformer with DINO~\citep{dino}, a vanilla self-supervised framework, which optimizes the similarities between two views through just one image-level representation per view. This is compared to pre-training with our proposed \modelname~and \textit{Cellback} frameworks. 
The encoders in our frameworks are implemented with both ViT-Tiny (ViT-T, $d=192$) and ViT-Small (ViT-S, $d=384$) consisting of 5M and 22M parameters, respectively.

\textbf{Dataset and tasks:} For \textit{slide-level classification}, we use the following datasets: 1) TCGA-Lung~\citep{tcga_luad, tcga_lusc} at 5$\times$, 2) TCGA-BRCA~\citep{tcga_brca} at 5$\times$, and 3) BRIGHT~\citep{BRIGHT}  at 10$\times$. Note that our proposed 
pre-training is performed separately for each dataset followed by evaluating them for slide-level classification. 
This classification task comprises 1) 
TCGA-Lung: Lung Adenocarcinoma (LUAD) versus Lung Squamous Carcinoma (LUSC), 2) TCGA-BRCA: Invasive Ductal (IDC) verses Invasive Lobular Breast Carcinoma (ILC), and 3) two sub-tasks in BRIGHT: 3-class WSI-classification (noncancerous, precancerous, and cancerous), and 6-class WSI-classification, termed as BRIGHT (3) and BRIGHT (6), respectively. 

 \noindent For \textit{patch-level classification}, evaluations are performed on Chaoyang \citep{chaoyang} and MHIST \citep{mhist} datasets, which contain localized annotation at crop-level. MHIST consists of two classes of colon cancer, whereas Chaoyang contains four classes of colon cancer. 

Note that, for generating cell prior $P_c$, we employed HoVer-Net for TCGA-Lung~\citep{hovernet} and due to computational limitations we employed Cellpose~\citep{cellpose} for the other three WSI datasets. 

\subsection{Datasets split}
\textbf{WSI Datasets:} 
1) \textit{TCGA-Lung}: This dataset consists of 940 diagnostic digital slides from two subtypes of Lung cancer - Lung adenocarcinoma (LUAD) and Lung Squamous cell carcinoma (LUSC). We split the data into 748 training (391 LUAD, 357 LUSC) and 192 (96 LUAD, 96 LUSC) testing samples randomly. The WSI-crops (670K train, 150K test) are extracted at 5$\times$ magnification. 
2) \textit{TCGA-BRCA}: This dataset consists 1034 diagnostic digital slides of two subtypes of Breast cancer - Invasive ductal carcinoma (IDC) and Invasive lobular carcinoma (ILC). We split the data into 937 training (747 IDC, 190 ILC) and 97 (77 IDC, 20 ILC) testing samples randomly. The WSI-crops (790K train, 90K test) are extracted at 5$\times$ magnification.
3) \textit{BRIGHT}: Comprises 703 (423 training, 80 validation, 200 testing) diagnostic digital slides. This dataset contains two sub-tasks: 3-class WSI classification and 6-class WSI classification tasks. For the first sub-task, the 3 classes are as follows - Non cancerous (PB+UDH), Pre-cancerous or Atypical (ADH+FEA), and Cancerous (DCIS+IC). For the second sub-task, the 6 classes are as follows - Pathological Benign (PB), Usual Ductal Hyperplasia (UDH), Flat Epithelia Atypia (FEA), Atypical Ductal Hyperplasia (ADH), Ductal Carcinoma in Situ (DCIS), and Invasive Carcinoma (IC). The BRIGHT challenge contains train, validation, and test splits. Since the challenge is not active now, labels for the test set are not available. Therefore, we reported our results for this dataset on its validation set as our test set. Class-wise data split can be found here \footnote{\href{https://research.ibm.com/haifa/Workshops/BRIGHT/}{https://research.ibm.com/haifa/Workshops/BRIGHT/}}. The WSI-crops (1.24M train, 0.2M test) are extracted at 10$\times$ magnification.

 \textbf{Patch Datasets:} 1) \textit{MHIST} \citep{mhist}: Consists of 3152 images of colon with tasks to classify the type of colorectal polyps into two types, benign and pre-cancerous. All the image resolutions are of 224 $\times$ 224 pixels. 2) \textit{Chaoyang} \citep{chaoyang}: Consists of 6160 patches of size 512 $\times$ 512 pixels from Colon cancer divided into four classes - normal, serrated, adenocarcinoma, and adenoma. These patches are resized to 224 $\times$ 224 pixels in our experiments. 
     
For these two patch datasets, we split their official training sets into a 5 fold cross validation sets. 
We train on 4 folds, validate on 1 fold and test on their official test sets. 
Thus, we report our results (accuracy and AUC) as a mean of 5 fold cross validation trials. 

\begin{table*}[ht]
\caption{Results for slide-level classification tasks. T denotes ViT-Tiny, and S denotes ViT-Small.}
\label{results}

\begin{center}
\begin{tabular}{ccccccccccc}
\toprule
\multicolumn{1}{c}{Dataset}                                             & \multicolumn{1}{c}{\textbf{Lung}}            & \multicolumn{1}{c}{\textbf{BRCA}}                                   & \multicolumn{1}{c}{\textbf{BRIGHT (3)}}       & \multicolumn{1}{c}{\textbf{BRIGHT (6)}}       \\ 
\multicolumn{1}{c}{Metric}                                             & \multicolumn{1}{c}{Acc, AUC}            & \multicolumn{1}{c}{Acc, AUC}                     & \multicolumn{1}{c}{Acc, AUC}            & \multicolumn{1}{c}{Acc, AUC}            \\
\midrule
Baseline-T                                                           & \multicolumn{1}{c}{0.894, 0.960}          & \multicolumn{1}{c}{0.907, 0.945}                 & \multicolumn{1}{c}{0.632, 0.850}          & \multicolumn{1}{c}{0.474, 0.776}          \\
Cellback-T                                                         & \multicolumn{1}{c}{\textbf{0.908}, \textbf{0.965}}          & \multicolumn{1}{c}{0.897, 0.940}          & \multicolumn{1}{c}{0.646, 0.848} & \multicolumn{1}{c}{0.488, 0.769} \\ 
\begin{tabular}[c]{@{}c@{}}
\modelnameplain-T \end{tabular} & \multicolumn{1}{c}{0.897, 0.957} & \multicolumn{1}{c}{\textbf{0.927}, \textbf{0.963}}         & \multicolumn{1}{c}{\textbf{0.653}, \textbf{0.852}}          & \multicolumn{1}{c}{\textbf{0.500}, \textbf{0.780}}          \\
\midrule

Baseline-S                                                           & \multicolumn{1}{c}{0.913, 0.967}          & \multicolumn{1}{c}{0.907, 0.947}               & \multicolumn{1}{c}{0.630, 0.840}          & \multicolumn{1}{c}{0.474, 0.781}          \\ 
Cellback-S                                                     & \multicolumn{1}{c}{\textbf{0.922}, 0.967}          & \multicolumn{1}{c}{\textbf{0.928}, 0.957}         & \multicolumn{1}{c}{0.667, 0.848} & \multicolumn{1}{c}{\textbf{0.529}, 0.796} \\ 

Cellback V2-S                                                         & \multicolumn{1}{c}{0.912,  \textbf{0.971}}          & \multicolumn{1}{c}{\textbf{0.928}, 0.958}               & \multicolumn{1}{c}{\textbf{0.692}, \textbf{0.861}}          & \multicolumn{1}{c}{\textbf{0.529}, 0.799}          \\

\begin{tabular}[c]{@{}c@{}}
\modelnameplain-S \end{tabular} & \multicolumn{1}{c}{0.911, \textbf{0.971}} & \multicolumn{1}{c}{\textbf{0.928}, 0.963}           & \multicolumn{1}{c}{0.662, 0.839}          & \multicolumn{1}{c}{\textbf{0.529}, \textbf{0.811}}          \\

\addlinespace[2pt]
\cdashline{2-5}
\addlinespace[2pt]
IN-S                                                           & \multicolumn{1}{c}{0.826, 0.908}          & \multicolumn{1}{c}{0.886, 0.953}               & \multicolumn{1}{c}{0.586, 0.766}          & \multicolumn{1}{c}{0.423, 0.712}          \\ 
CTransPath                                                          & \multicolumn{1}{c}{0.897, 0.956}          & \multicolumn{1}{c}{0.918, \textbf{0.971}}         & \multicolumn{1}{c}{0.632, 0.832} & \multicolumn{1}{c}{0.435, 0.757} \\ 
\begin{tabular}[c]{@{}c@{}}
RetCCL\end{tabular} & \multicolumn{1}{c}{0.870, 0.930} & \multicolumn{1}{c}{\textbf{0.928}, 0.962}           & \multicolumn{1}{c}{0.512, 0.852}          & \multicolumn{1}{c}{0.337, 0.682}          \\

\bottomrule
\end{tabular}
\end{center}
\end{table*}

\subsection{Implementation details.} 
For all our experiments, $224 \times 224$ sized crops are extracted from WSIs. We set the patch size for vision transformer input to $p=16$.
Therefore, the number of tokens per WSI crop are $n=196$. For ViT-Tiny (ViT-T), the embedding dimension $d=192$, whereas for ViT-small (ViT-S), $d=384$.

\textit{Pre-training:} For pre-training with DINO, we follow the hyper-parameter initialization from their source code~\citep{dino}. We use a batch size of 256.
In pre-training \modelname, we set the loss weighting factors $\lambda_1 = 0.5$ and $\lambda_2 = \frac{0.1}{4}$, whereas for \modelname~ without disentangle block (\textit{Cellback}), we set $\lambda_1 = 0.5$. We use two different projection heads of default output sizes (65536) \citep{dino} for region representations $f_c$ and $f_b$, and four different projection heads of smaller output size (4096) for disentangled spatial interplay features. For pre-training with i) SimCLR, we adopted the implementation from \citet{DSMIL} with a batch size 512, ii) EsViT, implementation is adopted from \citet{esvit}, and iii) SelfPatch, implementation is adopted from \citet{selfpatch}. Note that
pre-training is only performed on training samples of WSI datasets. For slide-level classification tasks, all models are  for 100 epochs on Lung dataset, for 50 epochs on BRIGHT, and 30 epochs on BRCA. To study data-efficiency plot in Fig. \ref{fig:dataefficiency}, all the models are  for 50 epochs on the Lung dataset.

\textit{Multiple instance learning:} We use DSMIL \citep{DSMIL} for slide-level classification throughout this study. 
 For training DSMIL, we use a learning rate of $2e^{-4}$, and weight decay of $5e^{-2}$. Batch size is set to 1 to handle variable bag size for each WSI $\mathcal{W}$. For each WSI-level MIL experiment, we run DSMIL with 10 different seeds and report the average performance. Other hyper-parameters such as number of epochs, and train-validation split ratio are kept consistent with \citet{DSMIL}. Note that training samples from the WSI datasets are split into train and validation for MIL training. 
 We hope to explore the impact of other MIL frameworks such as \citet{ABMIL, Transmil, CLAM, sparseconvmil, hipt, zhang2022gigapixel, pinckaers2020streaming} on our \modelname~learned features in future.

\textit{Patch classification:}
    In our experiments of fine-tuning for patch classification, an average pooling layer (for averaging the tokens) followed by a fully connected layer is placed on top of the 
     transformer-encoder backbone.
    For all experiments on MHIST dataset, we use a learning rate of $3e^{-4}$, weight decay of $1e^{-2}$, and batch size of 128. We train the network for 40 epochs and decay the learning rate by 0.1 at epoch 20 and epoch 30. 
    For all experiments on the Chaoyang dataset, we use a learning rate of $1e^{-4}$, weight decay of $1e^{-2}$, and batch size of 128. We train the network for 45 epochs and decay the learning rate by 0.1 at epoch 20, 30 and 40.

\subsection{Slide-level and Patch-level classification} \label{results_text}

\textbf{Slide-level classification:} In Table \ref{results}, we show the slide-level classification results on the three datasets with tiny and small ViT models  using \textit{Baseline}, \textit{Cellback} (\modelname~without disentangle block), and \modelname~frameworks. For state-of-the-art comparison, we employed pan-cancer pretrained models provided by CTransPath~\citep{wang2022transformer} and RetCCL~\citep{wang2023retccl} as the feature extractor instead of our pretrained models. For feature set extracted from both these models, we find that weight decay of $5e^{-3}$ in DSMIL works best. We have also included ImageNet pretrained ViT-Small provided by ~\citep{deit} as a feature extractor for comparing it's performance on slide-level tasks with DSMIL.

It may be observed that for the Lung and BRCA dataset, \modelname~consistently surpasses the \textit{Baseline} (up to 6.9\% relative accuracy gain) and \textit{Cellback} (up to 3.3\% relative accuracy gain) models for both ViT-T and ViT-S architectures. In all cases, both \modelname~and \textit{Cellback} considerably outperform the vanilla-DINO \textit{Baseline} (accuracy and AUC). Interestingly, \textit{\modelnameplain-T} performs even better than \textit{Baseline-S} in BRCA, substantiating the importance of efficiently encoding diversified information even in smaller feature embedding ($d=192$) in \modelname-T against inefficiently and sparsely encoding into a larger feature embedding ($d=384$) in ViT-S. This paves the direction for efficiently encoding domain-information in smaller models.
In the BRIGHT dataset, it is observed that \textit{Cellback-S} achieves the best performance for both the sub-tasks. Additional comparisons with SOTA SSL methods are provided in the \ref{additional_results}.

\textbf{Patch-level classification:} 
For evaluating the generalizability of the learned representations, we use BRCA pretrained models and fine-tune them on MHIST~\citep{mhist} and Chaoyang~\citep{chaoyang} datasets (because of visual similarities between breast and colon cancers \citep{bremond1984breast}). 
In Table~\ref{table:result_downstream_cv}, we report the 5-fold cross validation accuracy and AUC on the official test set.
We observed that both our models (\textit{Cellback} and \modelname) outperform the \textit{Baseline} on the two datasets using ViT-T and ViT-S backbones.

\begin{table}[ht]

\caption{Results for crop-level classification tasks.}
\label{table:result_downstream_cv}
\begin{center}
\begin{tabular}{ccc}
\toprule
Dataset       & \multicolumn{1}{c}{\textbf{MHIST}}                           & \multicolumn{1}{c}{\textbf{Chaoyang}}                        \\ 
Metric        & \multicolumn{1}{c}{Acc, AUC}                 & \multicolumn{1}{c}{Acc, AUC}            \\
\midrule
Baseline-T    & \multicolumn{1}{c}{0.758, 0.854}          & \multicolumn{1}{c}{0.819, 0.942}          \\ 
Cellback-T     & \multicolumn{1}{c}{0.769, 0.864}          & \multicolumn{1}{c}{0.823, 0.942}          \\ 
\modelnameplain-T  & \multicolumn{1}{c}{\textbf{0.782}, \textbf{0.871}} & \multicolumn{1}{c}{\textbf{0.828}, \textbf{0.945}} \\
\midrule
Baseline-S    & \multicolumn{1}{c}{0.757, 0.844}          & \multicolumn{1}{c}{0.830, 0.946}          \\ 
Cellback-S      & \multicolumn{1}{c}{0.765, 0.852} & \multicolumn{1}{c}{0.831, \textbf{0.950}} \\ 
\modelnameplain-S  & \multicolumn{1}{c}{\textbf{0.770}, \textbf{0.857}}          & \multicolumn{1}{c}{\textbf{0.836},  \textbf{0.950}} \\
\bottomrule
\end{tabular}
\end{center}
\end{table}

\begin{table*}[ht]
\caption{Comparison of \modelname~with existing dense pre-training SSL methods (SelfPatch and EsViT). Statistical significance analysis is provided in the Table~\ref{t-test}.}

\label{dense_comparison}
\begin{center}
\begin{tabular}{ccccccccccc}
\toprule
Dataset               & \multicolumn{1}{c}{\textbf{Lung}}                   & \multicolumn{1}{c}{\textbf{BRCA}}                                  & \multicolumn{1}{c}{\textbf{BRIGHT (3)}}             & \multicolumn{1}{c}{\textbf{BRIGHT (6)}}             \\ 
        Metric       & \multicolumn{1}{c}{Acc, AUC}            & \multicolumn{1}{c}{Acc, AUC}                      & \multicolumn{1}{c}{Acc, AUC}            & \multicolumn{1}{c}{Acc, AUC}            \\
\midrule
Baseline-S                                                           & \multicolumn{1}{c}{0.913, 0.967}          & \multicolumn{1}{c}{0.907, 0.947}               & \multicolumn{1}{c}{0.630, 0.840}          & \multicolumn{1}{c}{0.474, 0.781}          \\ 
SelfPatch-S    & \multicolumn{1}{c}{0.719, 0.785}          & \multicolumn{1}{c}{0.897, 0.940}                 & \multicolumn{1}{c}{0.513, 0.684}          & \multicolumn{1}{c}{0.363, 0.636}          \\ 
EsViT-S        & \multicolumn{1}{c}{0.914, 0.967}          & \multicolumn{1}{c}{\textbf{0.928}, 0.953}               & \multicolumn{1}{c}{0.678, 0.859} & \multicolumn{1}{c}{0.528, 0.788}          \\ 
Cellback-S                                                          & \multicolumn{1}{c}{\textbf{0.922}, 0.967}          & \multicolumn{1}{c}{\textbf{0.928}, 0.957}         & \multicolumn{1}{c}{0.667, 0.848} & \multicolumn{1}{c}{\textbf{0.529}, 0.796} \\ 

Cellback V2-S                                                            & \multicolumn{1}{c}{0.912,  \textbf{0.971}}          & \multicolumn{1}{c}{\textbf{0.928}, 0.958}               & \multicolumn{1}{c}{\textbf{0.692}, \textbf{0.861}}          & \multicolumn{1}{c}{\textbf{0.529}, 0.799}          \\

\begin{tabular}[c]{@{}c@{}}
\modelnameplain-S \end{tabular} & \multicolumn{1}{c}{0.911, \textbf{0.971}} & \multicolumn{1}{c}{\textbf{0.928}, \textbf{0.963}}           & \multicolumn{1}{c}{0.662, 0.839}          & \multicolumn{1}{c}{\textbf{0.529}, \textbf{0.811}}          \\
\bottomrule
\end{tabular}
\end{center}
\end{table*}

Relative to the \textit{Baseline}, \modelname~improves the accuracy by $1.7-3.16\%$ on MHIST and $0.8-1.1\%$ on the Chaoyang dataset.
Similarly, it also improves the corresponding AUCs by $1.5-2\%$ and $0.3-0.4\%$, respectively.

\begin{figure*}[t]
    \centering
    \includegraphics[width=1\linewidth]{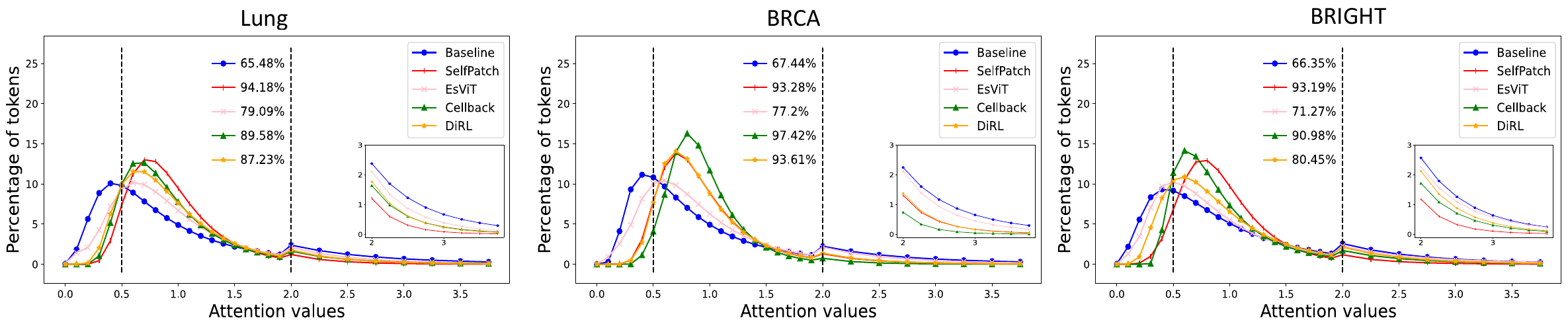}%
    \caption{\textbf{Attention distribution plot.} The second bin (0.5-2) is the desired one. Here, the percentage values show the fraction of tokens with attention values in the desired range. The baseline method has a higher fraction of tokens in lower range and higher range sparse bins, which is not ideal for digital pathology applications. }%
    \label{fig:attention_percent}%
\end{figure*}

\textbf{Comparison of \modelname~with other dense pre-training methods:} Note that our 
pre-training aligns with dense pre-training literature as we perform dense matching across two views through region-level and disentangled representations instead of just matching through one image-level representation.
For fair comparison, we re-implement dense pre-training methods closely related to our research: 1) SelfPatch \citep{selfpatch} and 2) EsViT \citep{esvit}. In addition to image-level matching as in DINO, SelfPatch enforces invariance against each patch/token and its neighbors, whereas EsViT enforces matching between all the corresponding patch-based tokens across views. Note that we use ViT-S as the encoder for both the techniques. In Table \ref{dense_comparison}, we showcase the results for SelfPatch and EsViT for slide-level classification tasks on all three datasets.

We find that our proposed models perform on par or better than EsViT on all the slide-level datasets. 
Whereas SelfPatch performs significantly worse in most tasks, possibly because neighborhood token invariance hardly exists in pathology images unlike for well-defined objects in natural images.
Thus, our proposed domain-inspired dense matching shows consistent improvements for slide-level classification compared to the other densely pretrained models.

\subsection{Analysis of learned Attention} \label{Analysis of learned Attention}
Here we demonstrate the de-sparsification of the learned attention of our \modelname~ pretrained models. 
Recall that the aggregated attention associated with a token is represented by the sum of all values across its corresponding column in the $n \times n$ self-attention matrix.

The sum of the aggregated attention values of all tokens ($n$) should be $n$. Due to this constraint, if the model attends to some tokens with high attention values, then the attention value associated with other tokens are reduced significantly. 
In Fig. \ref{fig:attention_percent}, we plot the distribution of aggregated attention values of tokens from the last layer of the transformer encoder pretrained by \textit{Baseline} (vanilla DINO), \textit{Cellback}, \modelname, EsViT, and SelfPatch. We then split the aggregated attention values in three bins: 0-0.5, 0.5-2, $>$2. The 0-0.5 and $>$2 bins indicate sparse attention learned through low and high concentrated attention values, respectively. Whereas the 0.5-2 range is the desired bin with moderate attention values that would lead to a de-sparsified attention map (and hence, optimal encoding of context-rich information).

The plots show that \textit{Baseline} trained with vanilla DINO has around 20-30\% tokens in the lower range sparse bin (0-0.5) and around 8-10\% in higher range sparse bin ($>$2). Whereas for our \textit{Cellback} and \modelname~models, fewer than 5-10\% tokens lie in the lower range sparse bin, while 3-5\% lie in higher range sparse bin.
Importantly, our models yield significantly more diversified attention with more than 80-90\% tokens in the desired bin (0.5-2) compared to that of 65\% tokens for the \textit{Baseline}. 
Interestingly, SelfPatch is able to diversify the transformer attention well, avoiding the sparse bins. However it still performs 2-20\% lower than our models on various slide-level classification tasks. This might be due to the neighbor invariant self-supervision (refer to \citet{selfpatch}) being noisy in histopathology domain (as discussed in \ref{results_text}). EsViT consistently contains 10-15\% more tokens compared to the \textit{Baseline} in the desired bin. However it still contains much more tokens in the sparse bins compared to \modelname~and \textit{Cellback}.
 These observations justify our premise that dense matching could diversify the attention, which is crucial for learning representations for histopathology.

\begin{figure*}[h]
    \centering
    \includegraphics[width=\linewidth]{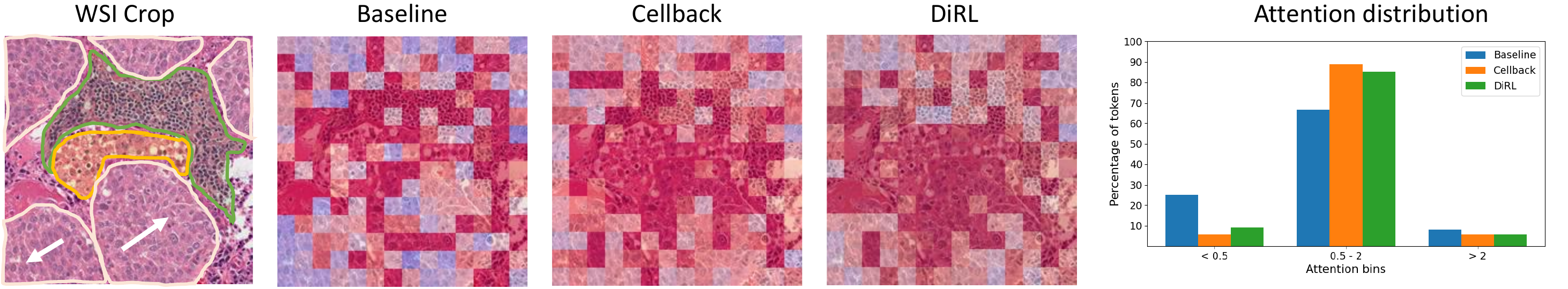}
    \caption{\textbf{Attention visualization.} Depicts the sparse attention by \textit{Baseline}, and its subsequent de-sparsification by our methods on a representative lung cancer patch. Bar plot shows the percentage of tokens in the three bins. \textit{Baseline} contains greater than 30\% of tokens' attention values in the sparse bins; comparatively our method contains fewer than 10\%.}
    \label{attention_visualization}
\end{figure*}

In Fig. \ref{attention_visualization}, we visualize the attention overlay from models pretrained using \textit{Baseline} vanilla DINO, and our proposed \modelname~and \textit{Cellback}. The regions containing tumor cells are outlined in white, while those with necrosis and immune cells are outlined in  yellow and green, respectively. It is evident that the \textit{Baseline} model is sparsely attending the WSI crop, often ignoring crucial tumor cell-dominant regions. In contrast, our models are able to globally diversify attention. Bar plots show that almost all tokens have moderate attention values ranging from 0.5 to 2 in \modelname. In contrast, \textit{Baseline} has a large number of tokens having very low attention ($<$0.5). Note that all the attention values $>1$ are clipped to 1 for visualization. Additional visualizations are provided in \ref{additional_attention_distribution_visualization}. 

\subsection{Ablation studies} \label{ablation}


Here we study the utility of various components proposed in our framework. We perform our ablations on the Lung cancer dataset.

\begin{figure}[ht]

  \begin{center}
    \includegraphics[width=0.8\linewidth]{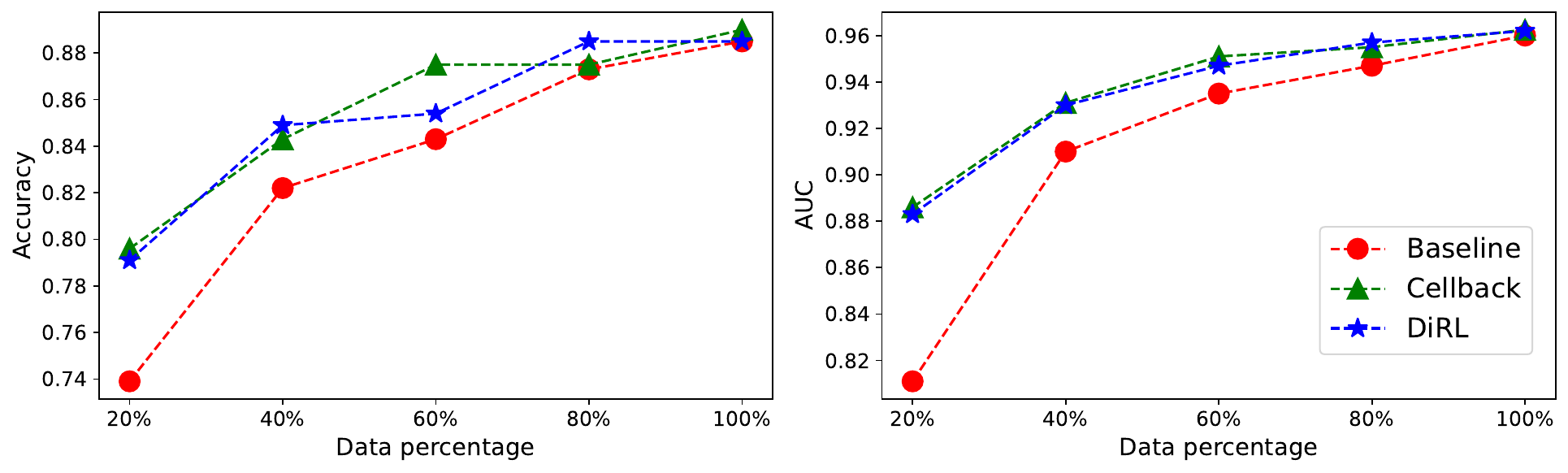}
  \end{center}
  \caption{\textbf{Data efficiency study.} Illustrates the effect of 
  pre-training with different amount of training data.}
  \label{fig:dataefficiency}
\end{figure}

\textbf{Data efficiency:} We  investigate the effect of pre-training with different fractions (20-100\%) of total training data. As seen in Fig. \ref{fig:dataefficiency}, the gain in both AUC and accuracy is around 6\% when \modelname-based models are pretrained with significantly less data (20\% data).  These empirical findings show the importance of \modelname~especially in low data regimes, for e.g. rare cancers.

\textbf{Adaptation of \modelname~with other SSL frameworks:}  In this study so far we adopted DINO framework for the self-supervision of the WSIs. However, \modelname~framework can be incorporated with any self-supervised learning strategies.
In Table \ref{simclr}, we demonstrate the performance of our proposed \textit{Cellback} and \textit{DiRL} representations pretrained with either BYOL \citep{byol} or SimCLR \citep{simclr} pipeline.
We compare this adapted framework with \textit{Baseline} models pretrained with vanilla BYOL and vanilla SimCLR. 
Irrespective of SSL framework, \textit{Cellback} and \textit{DiRL} consistently outperform the \textit{Baseline} in both Accuracy and AUC. 

\begin{table}[ht]
\caption{Pre-training \modelname~with other SSL frameworks. All the results are reported for models pretrained for 20 epochs.}
\label{simclr}
\begin{center}
\begin{tabular}{cccc}
\toprule
    \multirow{2}{*}{SSL framework}  &  Dataset       & \multicolumn{2}{c}{\textbf{Lung}}         \\ 
     &   Metric       & \multicolumn{1}{c}{Acc}       & AUC       \\  
\midrule
& Baseline-S     & \multicolumn{1}{c}{0.750}          &    0.821       \\ 
BYOL & Cellback-S     & \multicolumn{1}{c}{0.802}          &      \textbf{0.859}   \\ 
& DiRL-S    & \multicolumn{1}{c}{\textbf{0.812}}          &      0.854     \\ 

\midrule

& Baseline-S     & \multicolumn{1}{c}{0.791}          &    0.868       \\ 
SimCLR & Cellback-S     & \multicolumn{1}{c}{\textbf{0.805}}          &      \textbf{0.892}    \\ 
& DiRL-S     & \multicolumn{1}{c}{0.792}          &      0.891     \\ 

\midrule

& Baseline-S     & \multicolumn{1}{c}{0.808}          &    0.886       \\ 
DINO & Cellback-S    & \multicolumn{1}{c}{0.823}          &      0.903   \\ 
& DiRL-S     & \multicolumn{1}{c}{\textbf{0.825}}          &      \textbf{0.911 }    \\ 

\bottomrule

\end{tabular}
\end{center}
\end{table}

\textbf{Multi-task adaptation of vanilla SSL:} \label{multitask} An alternative approach of using cell segmentation as a prior could be to use the segmentation as an auxiliary task in self-supervised pre-training strategy. Consequently, in this section, we evaluate the impact of pre-training ViT with vanilla SSL jointly with a segmentation related auxiliary task. In order to avoid the use of a heavy decoder for segmentation, we instead design a new ‘cell prediction task’ to predict the number of cells present in each $p \times p$ patch of WSI-crop. A linear layer is applied on top of the ViT encoder for this task.

The joint optimization of vanilla SSL with the cell prediction task could hypothetically force the model to learn discriminative features from SSL and capture de-sparsification effects from the supervised loss ($L_{sup}$). We adopted the cell prediction supervised loss $L_{sup}$ with the \textit{Baseline}. 
We find that the model with $L_{sup}$ could outperform the baseline model at early epochs (accuracy of 0.832 vs. 0.808 in baseline, AUC of 0.921 vs. 0.886 in baseline). However, with later training epochs, the supervised loss does not augment the vanilla SSL pre-training (accuracy of 0.913 vs. 0.913 in baseline, AUC of 0.968 vs. 0.967 in baseline). Thus, the multi-task adaptation leads to better convergence at lower epochs but under-performs our proposed pre-training strategies when trained for a longer training schedule (100 epochs). 

\begin{figure}[ht]
    \centering
    \includegraphics[width=0.7\linewidth]{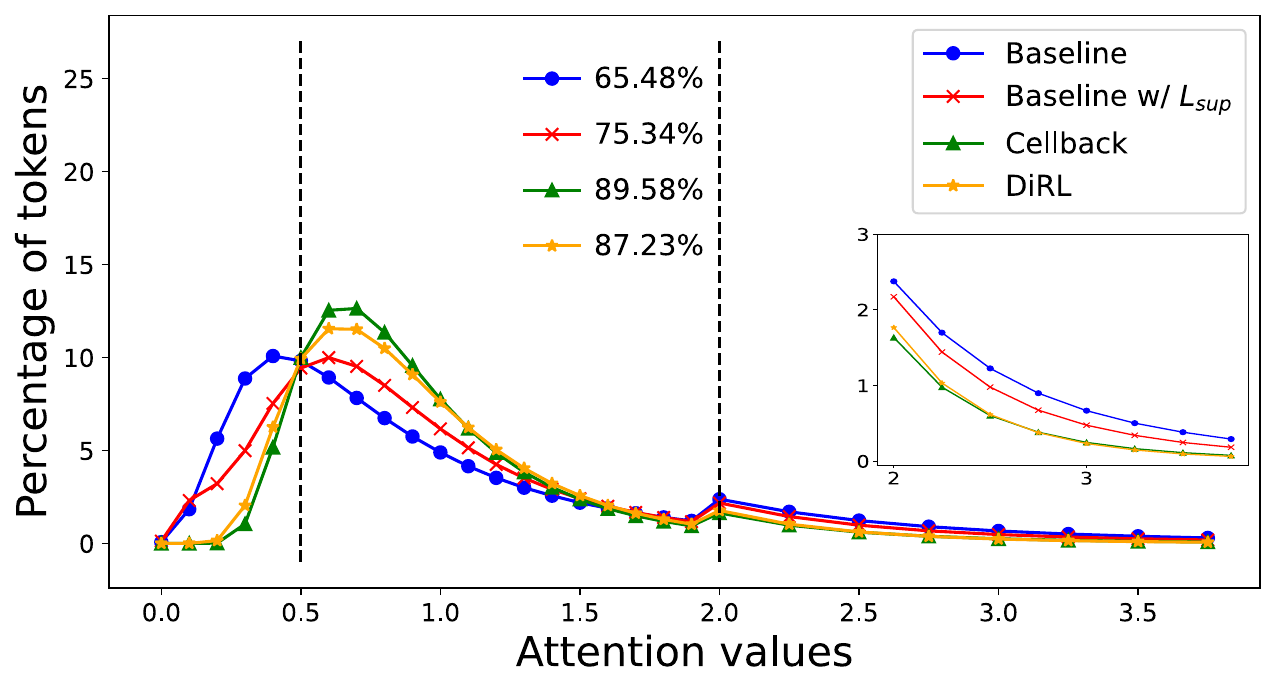}%
    \caption{Attention distribution plot. Vertical lines separate the three defined bins. The percentage values show the percentage of tokens with attention values in the desired bin (0.5-2) for \textit{Baseline}, \textit{Baseline w/ $L_{sup}$}, \textit{Cellback}, and \textit{DiRL}.}%
    \label{fig:attention_lung_multitask}%
\end{figure}

We analyzed the effect of $L_{sup}$ on the attention distribution in Fig. \ref{fig:attention_lung_multitask}. It reveals that although the supervised loss helps to de-sparsify attention to an extent, the de-sparsification is still sub-par compared to \textit{Cellback} and \textit{DiRL}. 

This ablation analysis shows that although cell prior could aid the representation learning with token-level supervision, the benefits are more prominent when leveraging cell-prior in our proposed dense pretext-task framework both in terms of performance and de-sparsification.

\begin{figure}[ht]
    \centering
    \includegraphics[width=0.7\linewidth]{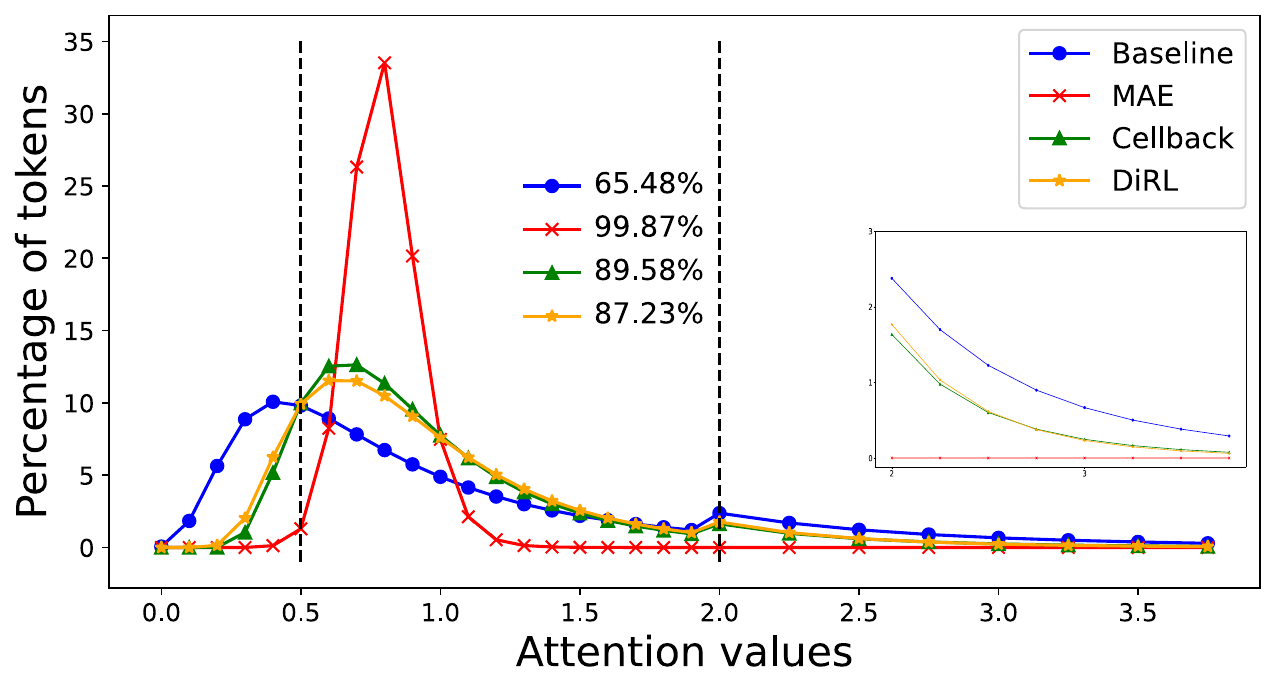}%
    \caption{Attention distribution plot. Vertical lines separate the three defined bins. The percentage values show the percentage of tokens with attention values in the desired bin (0.5-2) for \textit{Baseline}, \textit{MAE}, \textit{Cellback}, and \textit{DiRL}.}%
    \label{fig:attention_lung_mae}%
\end{figure}

\textbf{Comparison with Masked Image Modeling:} In this study, we hypothesized that the vanilla pretext task of matching the two views in SSL causes the sparsity in attention. Therefor,e a natural question that arises is that \textit{why not utilize the pretext task that explicitly focuses on local-level tasks such as reconstruction loss in Masked Autoencoders (MAE) \citep{mae}?} To investigate this, we pretrain the vision transformer with MAE-based loss for 200 epochs. We compare its slide-classification performance and attention de-sparsification in Table \ref{mae} and Fig. \ref{fig:attention_lung_mae} respectively. Our results reveal that although the attention de-sparsification is much better than other inter-view pretext tasks, the performance is significantly worse than all other methods. This is an expected observation since MAE is known to have subpar linear probing performance as the network isn't tasked to learn the discriminatory information. Since the pretrained network is directly utilized to extract the features for MIL, WSI-level tasks can be thought of as aligning to linear probing rather than fine-tuning. For more discussions on this phenomenon, we refer the reader to exemplary work in \citep{park2023self}. 

\begin{table}[ht]
\caption{Comparison with Masked Image Modeling}
\label{mae}
\begin{center}
\begin{tabular}{ccc}
\toprule
        Dataset       & \multicolumn{2}{c}{\textbf{Lung}}         \\ 
        Metric       & \multicolumn{1}{c}{Acc}       & AUC       \\  
\midrule
Baseline-S     & \multicolumn{1}{c}{0.913}          &    0.967       \\ 
MAE-S     & \multicolumn{1}{c}{0.820}          &      0.910   \\ 
Cellback-S     & \multicolumn{1}{c}{0.922}          &      0.967   \\ 
Cellback V2-S     & \multicolumn{1}{c}{0.912}          &      \textbf{0.971}   \\ 
DiRL-S     & \multicolumn{1}{c}{0.911}          &      \textbf{0.971}   \\ 
\bottomrule

\end{tabular}
\end{center}
\end{table}

Other ablations including: 1) \textit{Baseline} with an additional layer, 2) Effect of cell segmentation pipeline, and 3) Effect of stronger augmentation for diversification, are discussed in \ref{additional_ablation}.

\section{Discussion}

In this work, we present a crucial requirement of domain-driven tailoring of SSL techniques (proposed in natural imaging literature) to digital pathology tasks through our insightful observation about the sparsity of attention. We argued that since various natural imaging datasets are object centric, the sparsity in attention does not have an adverse effect on encoding capabilities particularly for global/shallow tasks like classification \citep{selfpatch}. However unlike object-centric natural images, pathology images are rather a complex phenotype of various spatially intermixed biological components. Therefore sparsity in attention leads to suboptimal encoding of this complex layout and thus could result in crucial information loss (see \ref{fig:additional_attention_visualization}). To address this critical unmet need, we proposed \modelname, a framework that densely encodes various characteristics of regions and their co-occurrences in pathology by leveraging region prior from cell segmentation. The proposed prior-guided pre-training utilizes densely extracted representations via a dense matching objective. We hypothesized that our pre-training strategy would make the attention more globally distributed, since matching multiple histopathology-specific representations would enforce the model to pay adequate attention to \textit{each} image-region relevant for corresponding characteristics.

Through our thorough qualitative analysis, we showed that \modelname~de-sparsifies the attention map, thus boosting the capabilities to encode diverse information in complex histopathology imaging. We believe that this attention diversification leads to a more ``complete" encoding of pathological components in WSI-crops. This was corroborated by consistent performance improvement on multiple slide-level and patch-level classification tasks by \modelname. We believe our work has the potential to augment future research in self-supervision/pre-training of vision-transformers for digital pathology domain. Our work opens exciting avenues toward utilizing domain-specific priors and instilling this domain knowledge in neural networks.

Limitation of \modelname~is although it gracefully leverages the flexibility available in vision-transformers, this approach cannot be trivially employed in CNNs. Since ViTs are computationally intensive, this limits the scope of our method in edge computing or low-resource environments. In future work we will explore more fine-grained concepts based on characteristics of tumor-immune microenvironment \citep{fassler2022spatial, abousamra2022deep, pati2020hact} to augment the priors.



\section*{Declaration of competing interest}
The authors do not have any competing interest.

\section*{Data availability}
Both slide-level and patch-level datasets are publicly available. Slide-level datasets are hosted on TCGA portal, whereas patch-level datasets are hosted on corresponding datasets' webpage. 

\section*{Acknowledgments}

Reported research was supported by the OVPR seed grant and ProFund grant at Stony Brook University, and NIH
1R21CA258493-01A1. The content is solely the responsibility of the authors and does not necessarily represent the official views of the National Institutes of Health. We also thank Maria Vakalopoulou and Ke Ma for their insights and valuable discussions.

\bibliography{conference}
\bibliographystyle{conference}

\newpage 

\appendix
\section{Appendix}

We provide additional details regarding the following in this section:
\begin{itemize}
    \item Additional experiments and results \ref{additional_results}
    \item Additional ablation studies \ref{additional_ablation}
    \item Additional attention visualization
    \ref{additional_attention_distribution_visualization}
    \item Disentangled block from pathological point of view
    \ref{disentangled_block}

\end{itemize}

\subsection{Additional experiments and results} \label{additional_results}

\begin{table*}[ht]
\caption{F1-score on BRIGHT validation set.}
\label{bright_challenge}
\begin{center}
\begin{tabular}{ccccc}
\toprule
\multicolumn{1}{c}{Dataset} & \multicolumn{1}{c}{\textbf{BRIGHT (3)}} & \multicolumn{1}{c}{\textbf{BRIGHT (6)}} & \multirow{2}{*}{\textbf{ROIs}} \\ 
\multicolumn{1}{c}{Metric}   & \multicolumn{1}{c}{F1-score}      & \multicolumn{1}{c}{F1-score}     \\  
\midrule
BRIGHT baseline \citep{brancati2021gigapixel}             & \multicolumn{1}{c}{0.580}   & \multicolumn{1}{c}{0.390}   &   \xmark \\
Method - 1 (SSL + CLAM)   \citep{wentai2022multiple}        & \multicolumn{1}{c}{0.642}     & \multicolumn{1}{c}{0.412}   & \xmark     \\ 
Baseline-S         & \multicolumn{1}{c}{0.589}        & \multicolumn{1}{c}{0.396}     & \xmark  \\ 
Cellback-S               & \multicolumn{1}{c}{0.644}      & \multicolumn{1}{c}{\textbf{0.450}}    & \xmark   \\ 
Cellback V2-S               & \multicolumn{1}{c}{\textbf{0.666}}      & \multicolumn{1}{c}{0.439}    & \xmark   \\ 
\modelnameplain-S          & \multicolumn{1}{c}{0.635}      & \multicolumn{1}{c}{0.446}    & \xmark   \\

\midrule

Method - 1  (FS + CLAM)  \citep{wentai2022multiple}        & \multicolumn{1}{c}{\textbf{0.691}}     & \multicolumn{1}{c}{\textbf{0.453}}   & \cmark    \\ 

Method - 2   \citep{zhan2022breast}         & \multicolumn{1}{c}{0.680}     & \multicolumn{1}{c}{0.440}    & \cmark    \\ 
Method - 3   \citep{marini2022multi}         & \multicolumn{1}{c}{0.650}     & \multicolumn{1}{c}{0.450}    & \cmark    \\ 
\bottomrule
\end{tabular}
\end{center}
\end{table*}

\textbf{Comparison with other methods in BRIGHT challenge on validation set:} 
Here we compare average F1-scores of our models with a few papers published as part of the BRIGHT challenge (\citet{wentai2022multiple,zhan2022breast,marini2022multi}), termed as Method - 1, Method - 2, and Method - 3 respectively, and the BRIGHT baseline presented by the challenge organizers in \citet{brancati2021gigapixel}. Note that BRIGHT challenge consists of WSI-level labels, as well as $3000 +$ annotated ROIs with an average size 2000 $\times$ 2000 pixels. Using them for supervision can naturally boost the performance. BRIGHT baseline doesn't use annotated ROIs. Method - 2 and Method - 3 utilized the ROIs for training their feature encoder. In contrast, Method - 1 presents two experiments, one with SSL on WSIs for feature encoder, and one with fully supervised training (FS) for training feature encoder. Following this, for MIL, they experimented with both CLAM and Reformer for slide-level classification. For MIL framework in our experiments we adopted DSMIL througout this study for slide-level tasks, and as DSMIL is more similar to CLAM than a transformer-based Reformer, for fair comparison we included their experiments with CLAM. It can be observed that for methods not supervised with localized annotation at ROI-level, \textit{Cellback V2} and \textit{Cellback}  performs best for 3-class and 6-class classification tasks respectively in terms of F1-score. Compared to supervised counterpart, our \textit{Cellback} still performs on-par for 6-class task.

\begin{table*}[ht]
\caption{Effect of additional layer in \textit{Baseline}}
\label{Extra_block_disentangled}
\begin{center}
\begin{tabular}{ccccccc}
\toprule
\multicolumn{1}{c}{Dataset} & \multicolumn{2}{c}{\textbf{Lung}} & \multicolumn{2}{c}{\textbf{BRIGHT (3)}} & \multicolumn{2}{c}{\textbf{BRIGHT (6)}} \\ 
\multicolumn{1}{c}{Metric}  & \multicolumn{1}{c}{Acc}      & AUC & \multicolumn{1}{c}{Acc}      & AUC      & \multicolumn{1}{c}{Acc}      & AUC      \\  
\midrule
Baseline-S      & \multicolumn{1}{c}{0.913}    & 0.967         & \multicolumn{1}{c}{0.630}    & 0.840    & \multicolumn{1}{c}{0.474}    & 0.781    \\ 
Baseline-S - 13      & \multicolumn{1}{c}{0.910}    & 0.964     & \multicolumn{1}{c}{0.661}    & 0.838 & \multicolumn{1}{c}{0.473}    & 0.777    \\ 
Baseline-S - 13*    & \multicolumn{1}{c}{0.913}    & 0.966      & \multicolumn{1}{c}{0.665}    & 0.854    & \multicolumn{1}{c}{0.478}    & 0.793    \\ 
Cellback-S        & \multicolumn{1}{c}{\textbf{0.922}}    & 0.967       & \multicolumn{1}{c}{0.667}    & 0.848    & \multicolumn{1}{c}{\textbf{0.529}}    & 0.796    \\ 

Cellback V2-S    & \multicolumn{1}{c}{0.912}   & \textbf{0.971}       & \multicolumn{1}{c}{\textbf{0.692}}    & \textbf{0.861}    & \multicolumn{1}{c}{\textbf{0.529}}    & 0.799    \\

\modelnameplain-S    & \multicolumn{1}{c}{0.911}    & \textbf{0.971}        & \multicolumn{1}{c}{0.662}    & 0.839    & \multicolumn{1}{c}{\textbf{0.529}}    & \textbf{0.811}    \\ 
\bottomrule
\end{tabular}
\end{center}
\end{table*}

\subsection{Additional ablation studies} \label{additional_ablation}

\textbf{\textit{Baseline} with an additional layer.} For a fair comparison with \modelname~pre-training which contains additional disentangle transformer block, we implement the \textit{Baseline} ViT-S with an additional layer, i.e. a model with 13 transformer blocks. We explored two versions of this model: 1) only one DINO projection head from the 13$^{th}$ layer (\textit{Baseline}-S -13), 2) two DINO projection heads, one from the 12$^{th}$ layer and another from the 13$^{th}$ layer (\textit{Baseline}-S -13$^*$). In Table \ref{Extra_block_disentangled}, both these models perform sub-par to our proposed 
pre-training, confirming the importance of our domain-aware design choice in \textit{Cellback} and \modelname~over just adding more model parameters.

\begin{figure*}[ht]
    \centering
    \includegraphics[width=\linewidth]{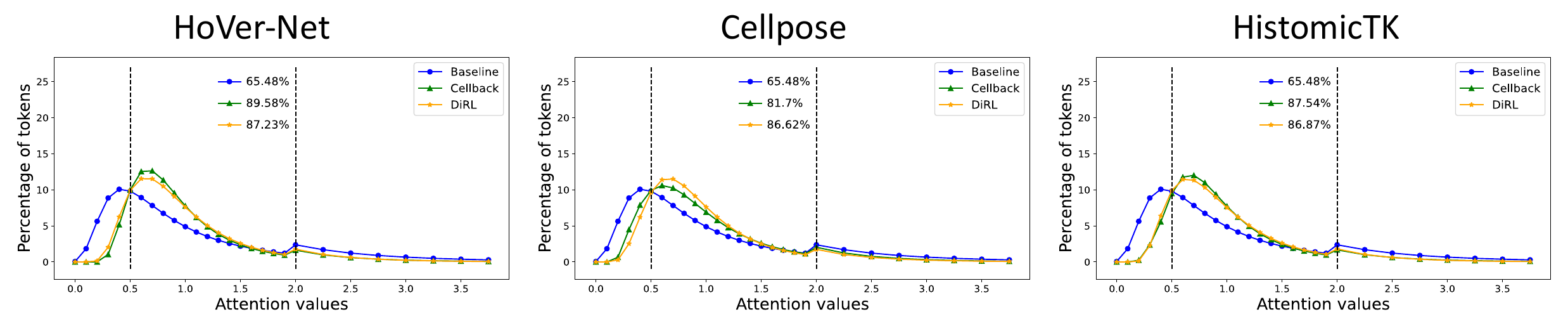}%
    \caption{Attention distribution plot. Vertical lines separate the three defined bins. The percentage values show the percentage of tokens with attention values in the desired bin (0.5-2) for \textit{Baseline}, \textit{Cellback}, and \textit{DiRL}.}%
    \label{fig:cell_segmentation_attention_effect}%
\end{figure*}

\begin{figure*}[ht]
    \centering
    \includegraphics[width=1\linewidth]{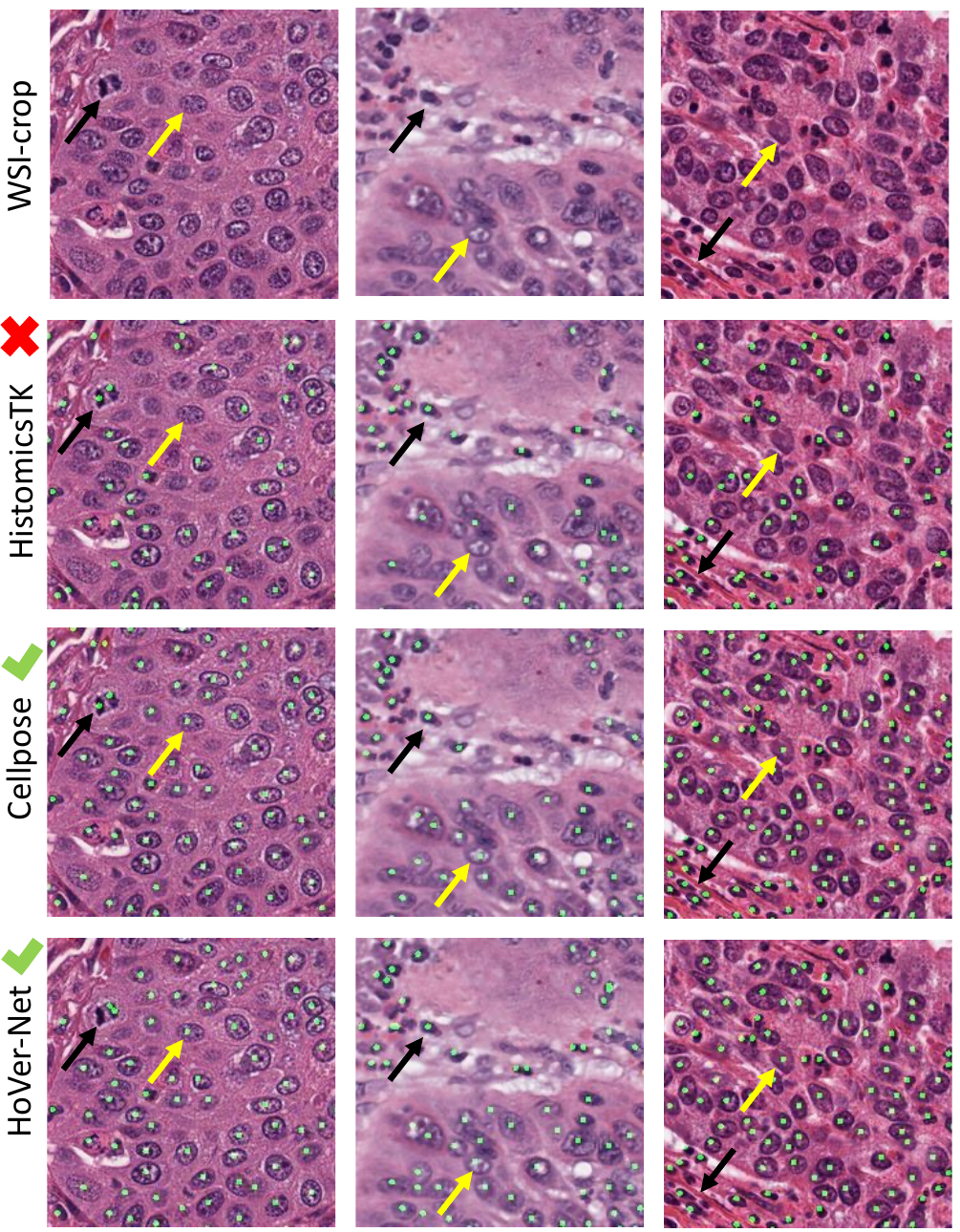}%
    \caption{\textbf{Cell segmentation performance.} Illustrations of WSI-crops, and the output of cell segmentation from HistomicsTK, Cellpose, and HoVer-Net. Black arrows denote normal cells/lymphocytes, and yellow arrows denote large tumorous cells. HistomicsTK misses numerous tumorous cells, while adequately detecting the normal ones. The more powerful deep-learning pipelines, Cellpose and HoVer-Net, are able to capture both type of cells with greater precision.}%
    \label{fig:cell_segmentation_3algo_4}%
\end{figure*}

\begin{figure*}[ht]
    \centering
    \includegraphics[width=1\linewidth]{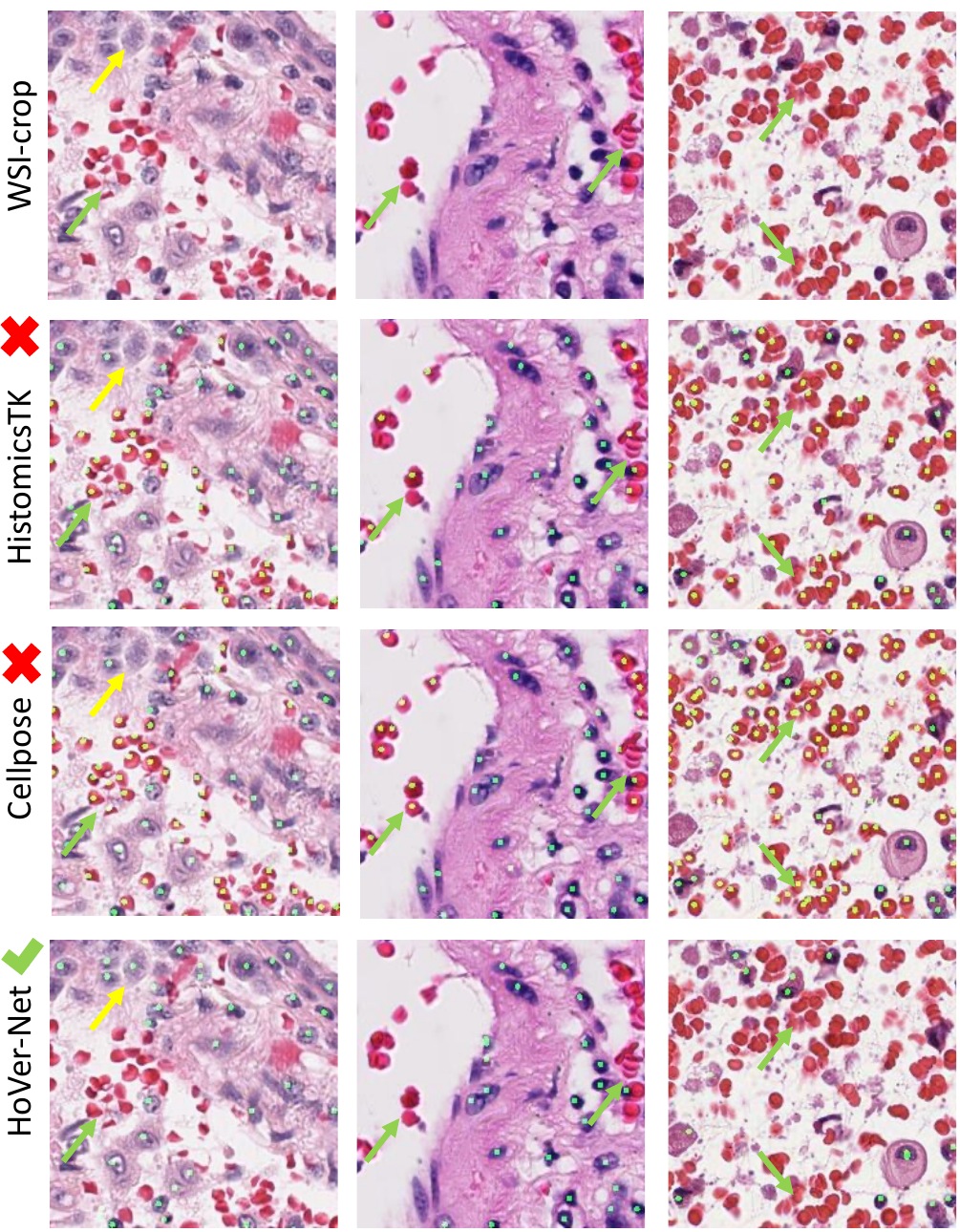}%
    \caption{Illustrations of WSI-crops, and the output of cell segmentation from HistomicsTK, Cellpose, and HoVer-Net. Green arrow denotes blood cells, and yellow arrow denotes large tumorous cells. Note that the arrangement/distribution of blood cells is not implicated in tumor phenotyping. Both HistomicsTK and Cellpose segment the blood cells, while HoVer-Net avoids them to a large extent. This example shows that HoVer-Net is preferable. 
    }%
    \label{fig:cell_segmentation_3algo_3}%
\end{figure*}

\textbf{Effect of cell segmentation pipelines:} We evaluate \modelname~framework with cell prior $P_c$ generated from three different cell segmentation pipelines namely HistomicsTK, Cellpose, and HoVer-Net on Lung cancer subtyping task in Table \ref{off_the_shelf}. HistomicTK is a python API which provides a handcrafted approach for cell segmentation based on~\citet{maxclustering,cdog,al2009improved}. The other two pipelines constitute powerful deep learning based models. Among the three, HoVer-Net $>$ Cellpose $>$ HistomicTK in terms of segmentation performance (see Fig. \ref{fig:cell_segmentation_3algo_4}, \ref{fig:cell_segmentation_3algo_3}). 

It can be observed that models trained with the near-perfect segmentation pipeline (HoVer-Net) performs the best in slide-level classification compared to HistomicTK and Cellpose. It is noteworthy that for other two segmentation pipelines, our method still performs on par with the vanilla DINO \textit{Baseline}, except for \textit{Cellback} model using HistomicTK for cell prior. This exception is attributed to the poor segmentation quality by HistomicTK, thus infusing noise in matching regions between the views in 
pre-training. We believe that using HoVer-Net on other datasets will further boost their performance compared to using Cellpose in Table \ref{results}. 
In Fig. \ref{fig:cell_segmentation_attention_effect}, we show the effect of cell segmentation pipelines on the de-sparsification of attention. Consistently across all the three segmentation pipelines, \textit{Cellback} and \textit{DiRL} achieve better de-sparsification compared to the \textit{Baseline}.


\begin{table*}[ht]
\caption{Effect of cell segmentation pipelines}
\label{off_the_shelf}
\begin{center}
\begin{tabular}{ccccccc}
\toprule
        Dataset      & \multicolumn{6}{c}{\textbf{Lung}}                                                                                                              \\ 
        Segmentation      & \multicolumn{2}{c}{HistomicsTK}                        & \multicolumn{2}{c}{HoVer-Net}                          & \multicolumn{2}{c}{Cellpose}      \\ 
        Metric      & \multicolumn{1}{c}{Acc}   & \multicolumn{1}{c}{AUC}   & \multicolumn{1}{c}{Acc}   & \multicolumn{1}{c}{AUC}   & \multicolumn{1}{c}{Acc}   & AUC   \\  
\midrule
Cellback-T    & \multicolumn{1}{c}{0.906} & \multicolumn{1}{c}{0.962} & \multicolumn{1}{c}{\textbf{0.908}} & \multicolumn{1}{c}{\textbf{0.965}} & \multicolumn{1}{c}{0.904} & 0.959 \\ 
\modelnameplain-T & \multicolumn{1}{c}{0.900} & \multicolumn{1}{c}{0.963} & \multicolumn{1}{c}{0.897} & \multicolumn{1}{c}{0.957} & \multicolumn{1}{c}{0.905} & 0.962 \\  
\midrule
Cellback-S    & \multicolumn{1}{c}{0.909} & \multicolumn{1}{c}{0.971} & \multicolumn{1}{c}{\textbf{0.922}} & \multicolumn{1}{c}{0.967} & \multicolumn{1}{c}{0.908} & 0.969 \\ 
\modelnameplain-S & \multicolumn{1}{c}{0.917} & \multicolumn{1}{c}{0.970} & \multicolumn{1}{c}{0.911} & \multicolumn{1}{c}{\textbf{0.971}} & \multicolumn{1}{c}{0.914} & \textbf{0.971} \\ 
\bottomrule
\end{tabular}
\end{center}

\end{table*}

\begin{figure}[ht]
    \centering
    \includegraphics[width=0.8\linewidth]{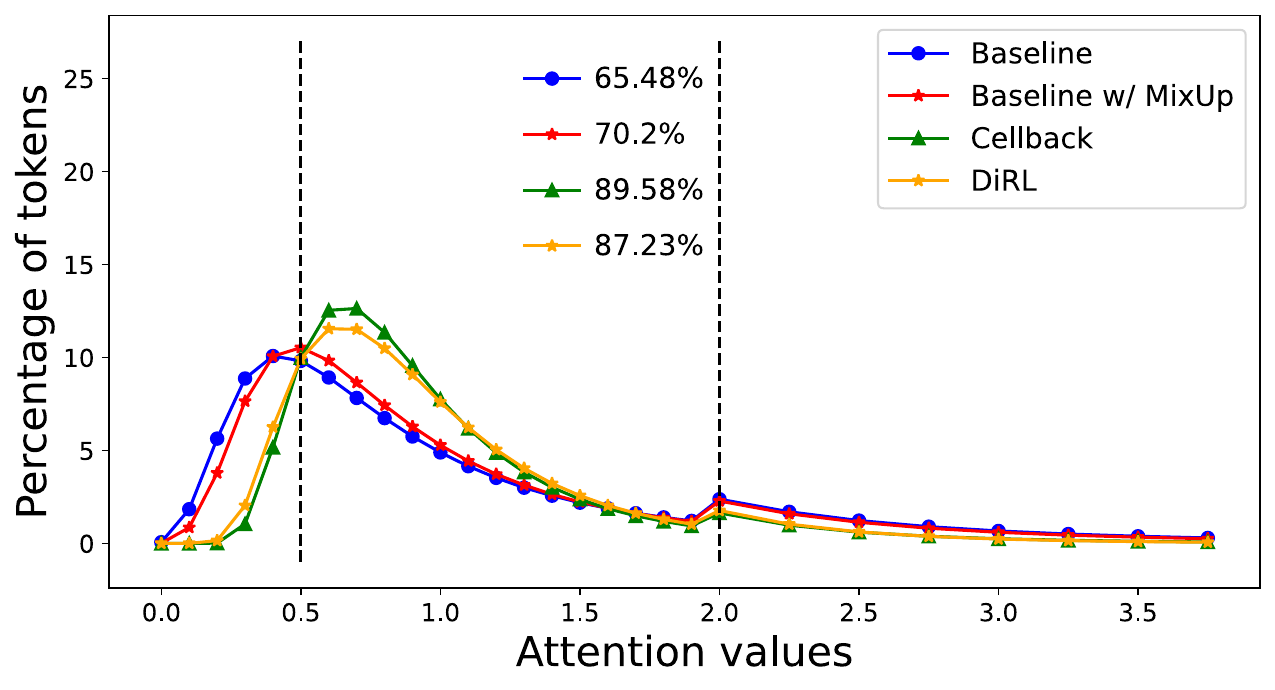}%
    \caption{Attention distribution plot. Vertical lines separate the three defined bins. The percentage values show the percentage of tokens with attention values in the desired bin (0.5-2) for \textit{Baseline}, \textit{Baseline w/ MixUp}, \textit{Cellback}, and \textit{DiRL}.}%
    \label{fig:attention_lung_mixup}%
\end{figure}

\textbf{Effect of MixUp in vanilla SSL:} Another alternative to diversifying model attention would be to use stronger augmentation techniques. To test this, we evaluate the effect of MixUp in pre-training a ViT with SSL. For applying MixUp in self-supervision, we adopted i-mix \citep{imix} in the DINO framework. In Table \ref{mixup}, it can be observed, that this MixUp strategy improves the performance of the vanilla SSL due to its regularizing effects. However, this improvement is not attributed to the de-sparsification of the transformer attention weights (see Fig. \ref{fig:attention_lung_mixup}). The complementary nature of our proposed approach and stronger augmentation techniques will be explored in future work. 

\begin{table}[ht]
\caption{Pre-training DINO with MixUp}
\label{mixup}
\begin{center}
\begin{tabular}{ccc}
\toprule
        Dataset       & \multicolumn{2}{c}{\textbf{Lung}}         \\ 
        Metric       & \multicolumn{1}{c}{Acc}       & AUC       \\  
\midrule
Baseline-S     & \multicolumn{1}{c}{0.913}          &    0.967       \\ 
Baseline-S w/ MixUp     & \multicolumn{1}{c}{0.916}          &      0.970   \\ 
Cellback-S     & \multicolumn{1}{c}{0.922}          &      0.967   \\ 
Cellback V2-S     & \multicolumn{1}{c}{0.912}          &      \textbf{0.971}   \\ 
DiRL-S     & \multicolumn{1}{c}{0.911}          &      \textbf{0.971}   \\ 
\bottomrule

\end{tabular}
\end{center}
\end{table}

\subsection{Additional attention visualizations} \label{additional_attention_distribution_visualization}

\begin{figure*}[ht]
    \centering
    \includegraphics[width=1\linewidth]{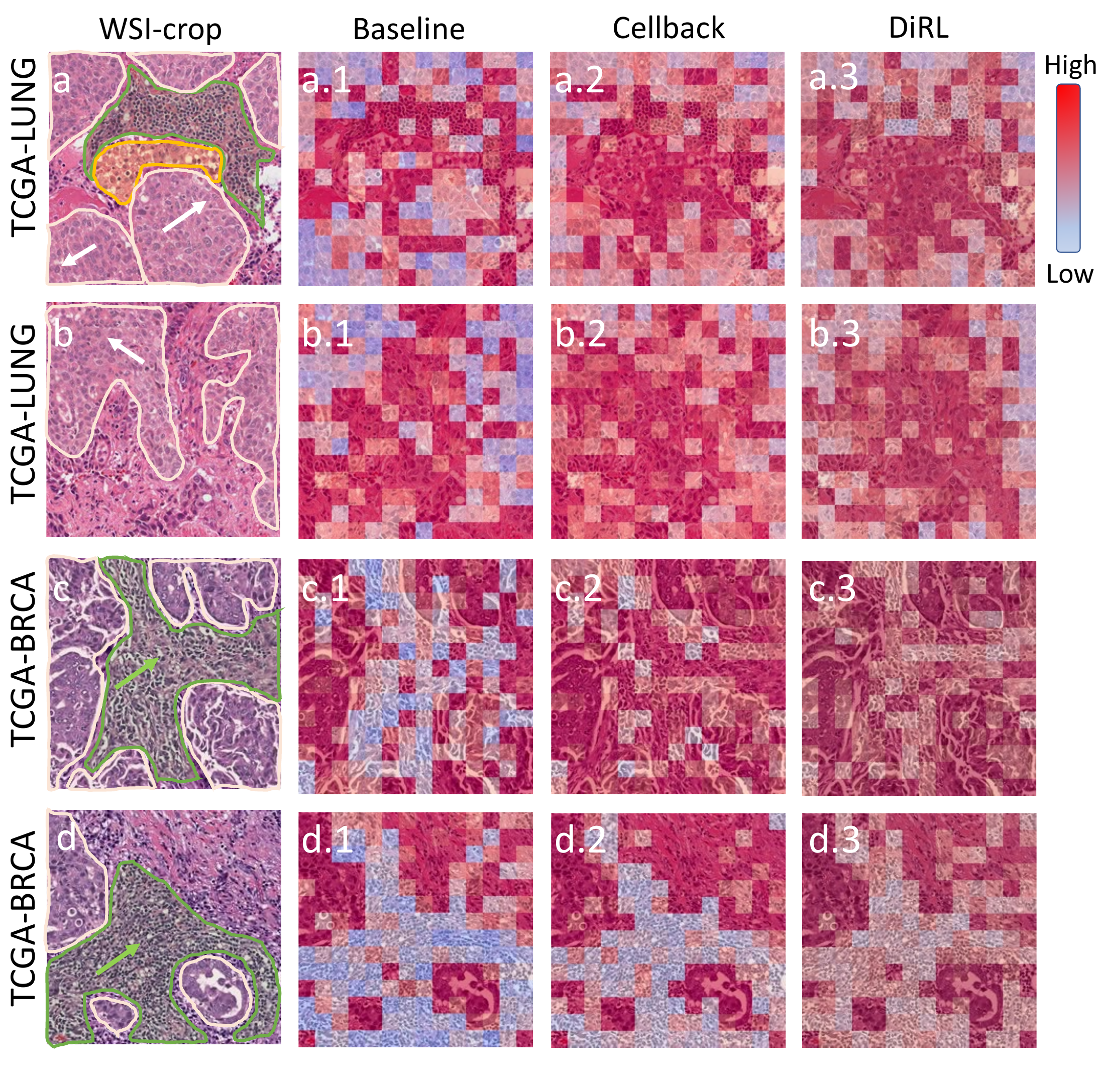}%
    \caption{\textbf{Attention visualization.} a and b shows an example 5$\times$ patches (size 224 $\times$ 224) from lung cancer, c and d shows an example 5$\times$ patch (size 224 $\times$ 224)  from breast carcinoma BRCA. In column 2, (*.1) shows the \textit{Baseline} model attention maps of the patches in column 1. In column 3, (*.2) shows the \textit{Cellback} model attention maps of the patches in column 1. In column 4, (*.3) shows the \modelname~attention maps of the patches in column 1. In a, b, c, and d, the tumor regions have been annotated in white, lymphocytic regions in green and necrosis in yellow. In the first and second rows, the arrows indicate the tumor regions which were sparsely attended to by the baseline as compared to our models. Though the baseline model attends well to the lymphocytic regions in a, it fails to densely attend to the important tumor areas. In the third and fourth row, the arrows indicate the lymphocytic regions which were sparsely attended to by the baseline as compared to our models. The tumor regions, however, exhibit high attention for all the models.
}%
    \label{fig:additional_attention_visualization}%
\end{figure*}

In Fig. \ref{fig:additional_attention_visualization}, we visualize the attention overlay from models pretrained using \textit{Baseline} vanilla DINO, and our proposed \modelname~and \textit{Cellback}. We mark various regions such as region dominated by tumor cell, immune cells. It can be observed that for Lung cancer, model predominantly focus on immune cell region, while ignoring much of the tumor cells region. Oppositely in BRCA samples, model attends to tumor cells region mainly, while ignoring the immune cells. This could result in loss of crucial information in encoding the representation by \textit{Baseline}. In contrast, \textit{Cellback} and \modelname~are able to adequately attend to each such region, thus encoding more informative representations.

\begin{figure*}[ht]
    \centering
    \includegraphics[width=1\linewidth]{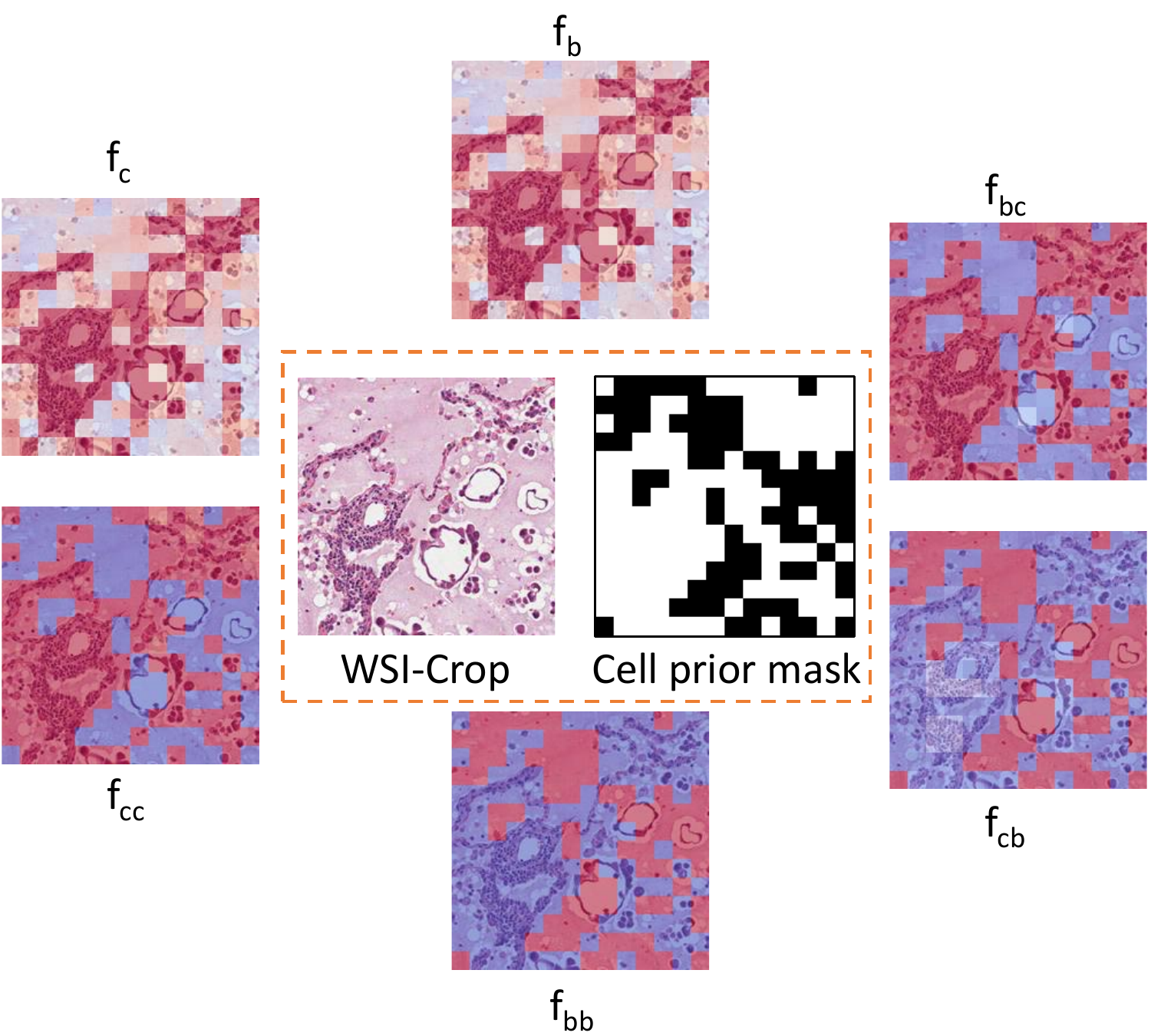}%
    \caption{\textbf{Representation wise attention maps.} 
}%
    \label{fig:attention_fc_fb_fcc_fbb_fcb_fbc}%
\end{figure*}

In Fig. \ref{fig:attention_fc_fb_fcc_fbb_fcb_fbc}, we investigate the attention map associated to region-level ($f_{c}$ and $f_{b}$) and disentangled representations ($f_{cc}$, $f_{bb}$, $f_{cb}$, and $f_{bc}$). For visualizing attention with respect to $f_{c}$ and $f_{b}$, we utilized the $n \times n$ self-attention matrix of last transformer block of the transformer encoder. The rows of this self-attention matrix corresponding to cell token indices are then selected and aggregated across the column to visualize the attention associated with cell tokens, thus describing attention associated with $f_{c}$. Similarly, background token indices are selected for attention visualization for $f_{b}$. For visualization with respect to disentangled representations, we use the $n \times n$ self-attention matrix from disentangle block, followed by adding it with $M_{self}$ and $M_{cross}$ separately. Thus we get two $n \times n$ matrices, one for $f_{cc}$ and $f_{bb}$, and the another for $f_{cb}$ and $f_{bc}$. This is followed by selecting cell/background rows and aggregation along column to get the attention values associated with each of the disentangled representation. Note that in all cases the attention is scaled by the corresponding number of cell/background tokens. This visualization shows that our modification in attention indeed works as expected, i.e. the masking ensures that cell tokens interact with just other cell tokens in $f_{cc}$ and similarly for background tokens in $f_{bb}$. Also, as expected, attention complementarity is observed between ($f_{cc}$ and $f_{bb}$) and ($f_{cb}$ and $f_{bc}$).

It can be observed that attention associated with cell token and background tokens in $f_{c}$ and $f_{b}$ are very similar. And same goes for ($f_{cc}$ and $f_{bc}$) and ($f_{cb}$ and $f_{bb}$). This is an known phenomenon~\citep{park2023self} that inter view-level pretext tasks makes the self-attentions collapse into homogeneity for all query tokens and heads. Therefore, even though different rows are selected from the $n \times n$ self-attention matrix, the scaled aggregated attention across column remains the same.

\subsection{Disentangled block from pathological point of view}
\label{disentangled_block}
In this study, we aim to model the interaction between the cellular (comprising various types of cells) and non-cellular regions (comprising stroma, smooth muscle region, fat, etc).
In pathology, the interaction between various entities (nuclei, stroma, glands, etc.) has been found to have clinical significance \citep{saltz2018spatial, diao2021human, zormpas2021superhistopath}. For modeling the interaction between cellular and non-cellular regions, we believe that disentangling both the regions followed by explicitly encoding the inter-intra region interaction is necessary. Otherwise  without disentanglement, encoding the inter-intra region interaction would not be precisely achievable, pertaining to the fact that average pooling in vision transformers could potentially dilute these token-level crucial signals. Future directions could delve into utilizing more refined entities (such as immune cells region, tumor regions, glands, necrotic region, and stroma) and quantifying their mutual interactions, thus better guiding the neural network to learn intricacies of digital pathology.


\newcolumntype{P}[1]{>{\centering\arraybackslash}p{#1}}

\begin{table*}[ht]

    \centering
    \caption{Statistical significance analysis (using t-test) between the results from 10 seeds from baseline and other dense methods (mentioned row-wise) compared with our proposed methods (mentioned column-wise) for various MIL tasks. p-values from the t-test are reported. ** denotes p-value$<$0.005, and * denotes p-value$<$0.05 while greater than 0.005. Takeaway: Compared to \textit{Baseline-S}, our methods (\textit{Cellback-S},\textit{ Cellback V2-S}, \textit{DiRL-S}) performed significantly better 10/12 times in terms of Acc and 9/12 times in terms of AUC. Compared with \textit{SelfPatch-S}: 12/12  (Acc) and 12/12 (AUC), and with \textit{EsViT-S}: 2/12  (Acc) and 7/12 (AUC).
}

    \begin{tabular}{c}
    \label{t-test}
        \textbf{Lung} \\
        \begin{tabular}{l|P{3.5cm}|P{3.5cm}|P{3.5cm}}
        \toprule
            & Cellback-S \newline Acc, AUC & Cellback V2-S \newline Acc, AUC & DiRL-S \newline Acc, AUC \\
            \midrule

         Baseline-S & $**$, $-$ & $-$, $**$ & $-$, $**$ \\
        SelfPatch-S & $**$, $**$ & $**$, $**$ & $**$, $**$ \\
        EsViT-S & $**$, $-$ & $-$, $**$ & $-$, $**$ \\

            \bottomrule
    \end{tabular} \\[1.5cm]
        
    \textbf{BRCA} \\
    \begin{tabular}{l|P{3.5cm}|P{3.5cm}|P{3.5cm}}
        \toprule
            & Cellback-S \newline Acc, AUC & Cellback V2-S \newline Acc, AUC & DiRL-S \newline Acc, AUC \\
            \midrule

        Baseline-S & $**$, $**$ & $**$, $**$ & $**$, $**$ \\
        SelfPatch-S & $**$, $**$ & $**$, $**$ & $**$, $**$ \\
        EsViT-S & $-$, $**$ & $-$, $**$ & $-$, $**$ \\
        
            \bottomrule
    \end{tabular} \\[1.5cm]
        
    \textbf{BRIGHT (3)} \\
    \begin{tabular}{l|P{3.5cm}|P{3.5cm}|P{3.5cm}}
        \toprule
            & Cellback-S \newline Acc, AUC & Cellback V2-S \newline Acc, AUC & DiRL-S \newline Acc, AUC \\
            \midrule

        Baseline-S & $**$, $*$ & $**$, $**$ & $**$, $-$ \\
        SelfPatch-S & $**$, $**$ & $**$, $**$ & $**$, $**$ \\
        EsViT-S & $-$, $-$ & $*$, $-$ & $-$, $-$ \\
        
            \bottomrule
    \end{tabular} \\[1.5cm]
        
    \textbf{BRIGHT (6)} \\
    \begin{tabular}{l|P{3.5cm}|P{3.5cm}|P{3.5cm}}
        \toprule
            & Cellback-S \newline Acc, AUC & Cellback V2-S \newline Acc, AUC & DiRL-S \newline Acc, AUC \\
            \midrule

        Baseline-S & $**$, $-$ & $**$, $**$ & $**$, $**$ \\
        SelfPatch-S & $**$, $**$ & $**$, $**$ & $**$, $**$ \\
        EsViT-S & $-$, $-$ & $-$, $*$ & $-$, $**$ \\
        
            \bottomrule
    \end{tabular}
    \end{tabular}
\end{table*}

\begin{figure*}[ht]
    \centering
    \includegraphics[width=1\linewidth]{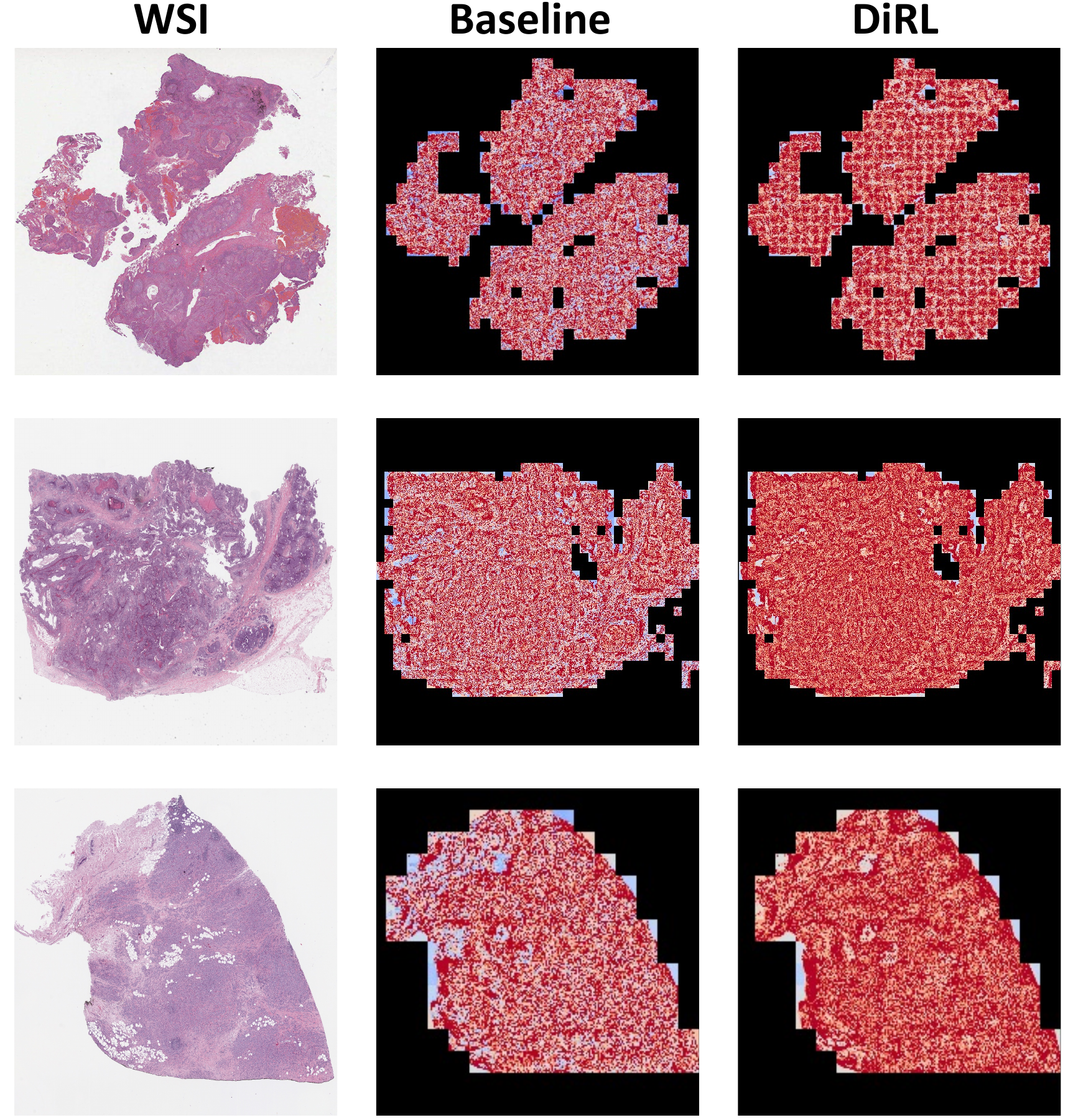}%
    \caption{\textbf{WSI-level self-attention visualization.}
}%
    \label{fig:wsi_level_attention_visualization}%
\end{figure*}

\begin{figure*}[ht]
    \centering
    \includegraphics[width=1\linewidth]{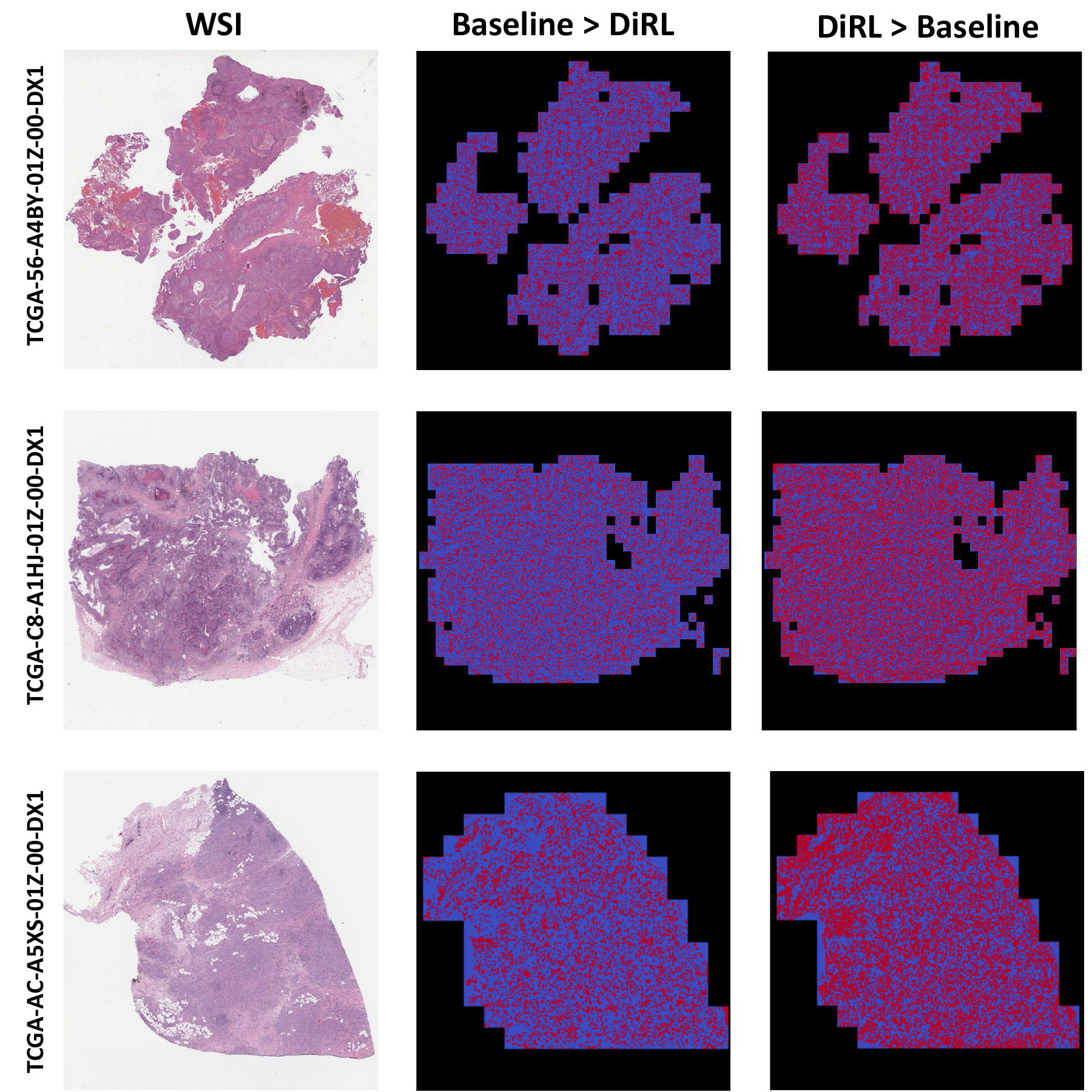}%
    \caption{\textbf{WSI-level self-attention difference visualization.} 
}%
    \label{fig:wsi_level_attention_difference_visualization}%
\end{figure*}

\end{document}